\documentclass{article}

\usepackage{arxiv}

\usepackage[utf8]{inputenc} %
\usepackage[T1]{fontenc}    %
\usepackage{hyperref}       %
\usepackage{url}            %
\usepackage{booktabs}       %
\usepackage{amsfonts}       %
\usepackage{nicefrac}       %
\usepackage{microtype}      %
\usepackage{lipsum}

\usepackage[section]{placeins}
\usepackage[shortlabels]{enumitem}
\usepackage{subcaption}
\usepackage{cite}
\usepackage{amsmath,amssymb,amsfonts}
\usepackage{algorithmic}
\usepackage{graphicx}
\usepackage{textcomp}
\usepackage{xcolor}
\def\BibTeX{{\rm B\kern-.05em{\sc i\kern-.025em b}\kern-.08em
    T\kern-.1667em\lower.7ex\hbox{E}\kern-.125emX}}
    
\usepackage{verbatim}  
    
\usepackage{hyperref}
\usepackage{csquotes}
\usepackage{todonotes}
\usepackage{float}

\usepackage{ tipa }

\newcommand{\terminal}{\textbf}
\newcommand{\nonterminal}{\text}

\title{A Journey in ESN and LSTM Visualisations\\on a Language Task}

\author{Alexandre Variengien$^{4,1,2,3}$,  and Xavier Hinaut$^{1,2,3,,*}$ 
\bigskip \\ 
\small
$^1$Inria Bordeaux Sud-Ouest, Talence, France\\
$^2$LaBRI, Université de Bordeaux, CNRS UMR 5800, Talence, France\\
$^3$IMN, Université de Bordeaux, CNRS UMR 5293, Bordeaux, France\\
$^4$Ecole Normale Supérieure de Lyon, 46 allée d’Italie, 69364 Lyon Cedex 07, France\\
$^*$Corresponding author: xavier.hinaut@inria.fr}

\date{\small

}

\begin{document}
\maketitle

\begin{abstract}
Echo States Networks (ESN) and Long-Short Term Memory networks (LSTM) are two popular architectures of Recurrent Neural Networks (RNN) to solve machine learning task involving sequential data. However, little have been done to compare their performances and their internal mechanisms %
on a common task. In this work, we trained ESNs and LSTMs on a Cross-Situationnal Learning (CSL) task. This task aims at modelling how infants learn language: they create associations between words and visual stimuli in order to extract meaning from words and sentences.
The results are of three kinds: performance comparison, internal dynamics analyses and visualization of latent space.
(1) We found that both models were able to successfully learn the task: the LSTM reached the lowest error for the basic corpus, but the ESN was quicker to train. Furthermore, the ESN was able to outperform LSTMs on datasets more challenging without any further tuning needed.
(2) We also conducted an analysis of the internal units activations of LSTMs and ESNs. %
Despite the deep differences between both models (trained or fixed internal weights), we were able to uncover similar inner mechanisms: 
both put emphasis on the units encoding aspects of the sentence structure.
(3) Moreover, we present \textit{Recurrent States Space Visualisations} (RSSviz), a method to visualize the structure of latent state space of RNNs, based on dimension reduction (using UMAP). %
This technique enables us to observe a fractal embedding of sequences in the LSTM. %
RSSviz is also useful for the analysis of ESNs (i) to spot difficult examples and (ii) to generate animated plots showing the evolution of activations across learning stages. %
Finally, we explore qualitatively how the RSSviz could provide an intuitive visualisation to understand the influence of hyperparameters on the reservoir dynamics prior to ESN training.

\end{abstract}

\keywords{ ESN \and LSTM \and Cross-Situational Learning \and Visualisation \and UMAP \and Dimension Reduction}

\section{Introduction}

Recurrent neural networks are now a mainstream architecture to deal with time-series or symbolic sequences. A wide variety of models have been proposed, among which, Long-Short Term Memory networks (LSTM) \cite{hochreiter1997long} are probably the most popular. They have been initially proposed as a solution to the vanishing gradient problem encountered when training simple RNNs with Back-propagation Through Time (BPTT) \cite{rumelhart1986learning}. To this end Hochreiter \& Schmidhuber used gating mechanisms to select which information to keep and which to forget, along with the ability to keep the gradient constant. This gating ability enables LSTMs to deal with sequences that present long-term dependencies.
In the past decade, they have been successfully applied to diverse problems, from traffic forecasting \cite{ma2015long}, to acoustic modeling \cite{sak2014long} and more famously, to natural language processing \cite{sundermeyer2012lstm}, a known difficult task because of the inherent long time dependencies. They often achieved state-of-the-art on different benchmarks. %

Echo State Networks is another approach to cope with the difficult training of RNNs. Instead of training the recurrent connections through Back-Propagation Through Time, the input and recurrent layer of ESNs are initialized at random and, according to the Reservoir Computing paradigm, are kept fixed during training\footnote{Some studies use unsupervised learning to adapt the weights inside the recurrent layer, such as homeostasic mechanisms like Intrinsic Plasticity \cite{steil2007online}}. Only the weights of the connections from the reservoir units to the output layer are learnt. It makes these output connections easier to train and is computationally less expensive.
Despite this simple architecture, ESNs have been successful on various task, from time series forecasting for wireless communication \cite{jaeger2004harnessing} to motor control \cite{salmen2005echo} or artificial grammar learning \cite{tong2007learning}. %

There have been some attempts at comparing both models, especially on time series prediction benchmarks, on which ESNs were found to outperform LSTMs \cite{gallicchio2018comparison, vlachas2020backpropagation}. 
As for language related tasks, some work have been done on Word Sense Disambiguation where a BiESN architecture were compared to a BiLSTM one \cite{popov2019echo}. The results obtained were comparable but the ESN architecture was much less computationally demanding. That is also the conclusion of \cite{jirak2020echo} in which classical ESNs and LSTMs were compared on a gesture classification task.

Even if both neural networks performs well, they are often seen as black-box models. Their internal dynamics remain difficult to understand because of the non-linear and high dimensional representations they deal with. 
Interesting work has been done to visualize such neural networks, especially for LSTMs. 
With a character prediction model trained on large corpora, Karpathy, Johnson and  Fei-Fei \cite{karpathy2015visualizing} investigated the activity of LSTM cells and discovered human interpretable activities.
For example, they found that some precise cells were active when the string of characters processed where inside quotation marks, and similarly others were sensitive to the length of a line in a paragraph. Since then, a lot of work as been done to make sense of the activity of inner cells of LSTMs trained as language models. Beyond being able to keep track of simple structure in a sentence such as shown by Karpathy, Johnson and  Fei-Fei, individual cells of LSTM can also encode  high level features such as positive or negative sentiment of a review, %
as shown in \cite{DBLP:journals/corr/RadfordJS17}. The tool \textit{LSTMvis} has been developed \cite{strobelt2017lstmvis} to easily visualize cell activation of LSTMs. Other approaches have been proposed such as in \cite{madsen2019visualizing} where, instead of visualizing cell activation, for different gated architectures they looked at the gradient of output activation (according to an input at a given time). %
It gives hindsights at which piece of information the LSTM pays attention when generating a given output.

Concerning ESNs however, few works have been done to explore the inner dynamics of reservoirs and represent how the trained output layer extracts information from reservoir states. In \cite{bianchi2016investigating}, the authors focused on a dynamical analysis of the reservoir for time-series. They used recurrence plots as a way to both quantify and visually characterize stability of the reservoir.
Of course, Jaeger proposed Conceptors as an extension to ESNs \cite{Jaeger2014, Jaeger2017} that enables to store patterns -- a conceptor stores the internal dynamics evoked by an input pattern -- and make logical operation with them. However, to our knowledge little work has been done to visualize and make sense of the internal dynamics of an ESN using recently developed visualisation methods, in particular the ones used in deep learning. At first sight, this is not surprising because internal ESN dynamics (in absence of feedback) are only determined by the inputs and the random weights: thus, these dynamics are unrelated to the desired tasks, one could just assume that they are \enquote{suitable} for a particular task. However, the projecting back output activations on internal dynamic representations can give interesting insights, for example to understand how the ESN performs a task, and why it fails to classify some input patterns.

In a more general way, there has been successful attempts to visualize high dimensional representations of artificial neural network. The recent dimensional reduction technique such as the Uniform Manifold Approximation and Projection (UMAP) \cite{mcinnes2018umap} enables detailed visualization in a 2D space of high dimensional structures. This tool has been successfully applied to understand the structure of inner representations of a Deep Convolutional Neural Network \cite{carter2019activation} where it enables users to spot adversarial examples. Concerning task related to language, UMAP was applied to the BERT model \cite{devlin2018bert} to understand the inner embedding of polysemous words \cite{coenen2019visualizing}.
Embedding visualisation by Principal Component Analysis (PCA) is used in the Embedding Comparator tool \cite{boggust2019embedding}. Dimension reduction can also be performed by an autoencoder, as done for hidden cells of LSTM trained for sign language recognition by Piotr Tempczyk and Mirosław Bartołd in this blog post \cite{sign-language-lstm}.
Concerning ESNs, UMAP has been anecdotally used in \cite{abdelrahman2019analyzing} to visualize the structure of the reservoir state on Lorentz series. 

To compare ESNs and LSTMs we chose a task that give insights on how human brains incrementally learn, from the perspective of a a child.
It is a psycholinguistic task called Cross-Situational Learning (CSL). It is inspired by developmental psychology experiments that aim to understand how infants learn the meaning of words in a fuzzy context: they need to make associations between symbols and referents (e.g. between words and objects). The learning extracts these associations from co-occurrences of words and stimuli. Recently, ESNs have been successfully applied to such task \cite{juven2020cross},
building on top of previous studies using fully supervised ESNs to model human sentence parsing \cite{dominey2006neurolinguistic, hinaut2013real}, multilingual processing \cite{hinaut2015recurrent, hinaut2019teach} and adapting it for human-robot interactions \cite{hinaut2014exploring, twiefel2016semantic, hinaut2018input}. 
To this extent, such cross-situational task is more plausible from a human brain learning perspective than a purely supervised task. This type of task is also interesting for practical applications were exact target outputs are not always available.

Similarly,  works have been done on language acquisition task with "LSTM-like" architectures (i.e. with gating mechanism). In \cite{zhong2017toward}, the authors used gated recurrent units (GRU) and variants to perform a learning robotic multi-modal task given sentences. In \cite{ororbia2018like},
augmenting predictive language models (based on gated architectures) with images enhances next-word prediction abilities. 
Moreover this task is interesting because the teacher outputs provide fuzzy information instead of the real desired outputs. This is due to the fact that the visual representation of a scene (e.g. several objects are visible by the infant) often contain more information than the utterances provided by the caregivers (e.g. \enquote{This is a spoon.}).

In this study, in section~\ref{sec-methods} we first quickly present ESNs and LSTMS, together with the Cross-Situational learning task used to compare both networks. This task is equivalent to a multi-class symbolic sequences classification, but which has the advantage of being human understandable. In section~\ref{sec-perf-results} we compare the performances of both networks on the task, and explore the influence of the size of the vocabulary on performance. In section~\ref{sec-comp-single-cell}, we investigate single unit activities in order to understand the inner mechanisms of both networks (like feature selection for ESNs and feature creation for LSTMs). %
In section~\ref{sec-RSS_ESN}, we present \textit{Recurrent States Space Visualisations} (RSSviz): a common framework to visualize different RNN based models on dimensional reduction of the recurrent states.  
In section~\ref{sec-pract-use-cases-RSS}, we present practical use-cases for RSSviz, including the influence of hyper-parameters on reservoir dynamics.
Finally, in section~\ref{sec-discussion} we discuss our quantitative and qualitative results along with perspectives.
In section~\ref{sec-sup-mat} we provide supplementary material along with the source code in Python available on Github.

\section{Methods}
\label{sec-methods}

\subsection{Echo State Networks and Reservoir Computing}

\subsubsection{Reservoir Computing}
\begin{figure}
    \centering
    \includegraphics[scale = 0.7]{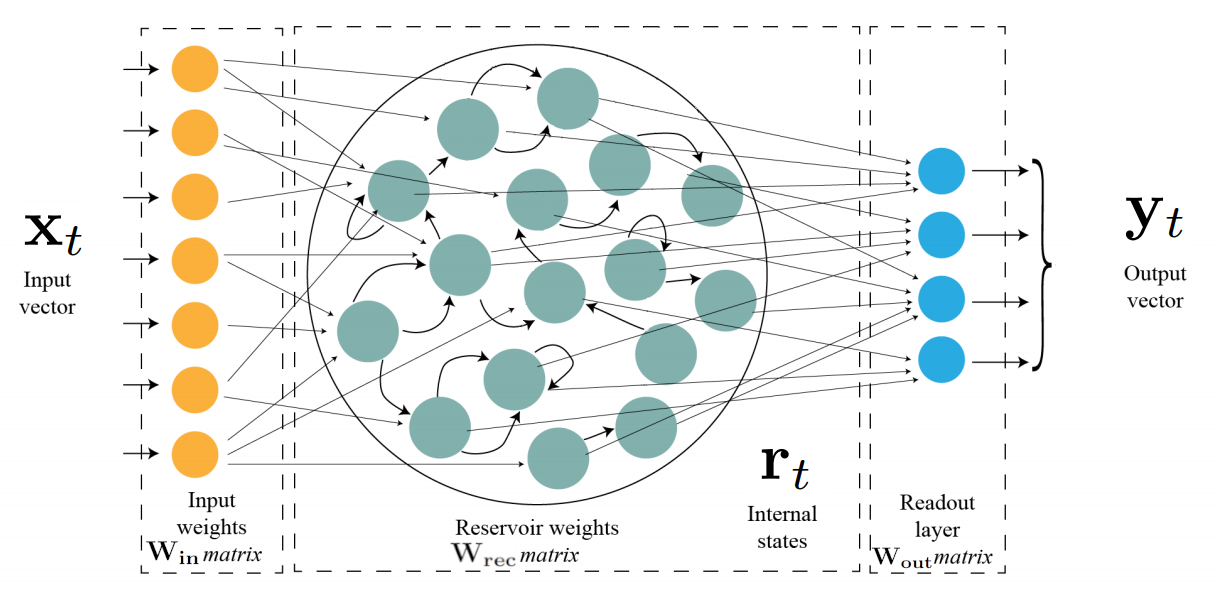}
    \caption{The ESN is trained to extract the target activation from the dynamic of the reservoir. To do this, only the readout connections, i.e. the coefficient of the matrix $\mathbf{W_{out}}$, are trained. This image was adapted from \cite{trouvain2020reservoirpy} with the authorisation of the authors.
    }
    \label{fig:res_schema}
\end{figure}

In this study, we worked with Echo States networks \cite{Jaeger2001}. They are part of the Reservoir Computing paradigm \cite{Lukoeviius2009}. They are a simple yet effective RNN model, fast and easy to train. Indeed, they keep a lot of weights untrained. The global architecture is depicted in figure \ref{fig:res_schema}.

The reservoir state $r_t$ is updated according to the equation \ref{eq_reservoir}. $\mathbf{W_{in}}$ corresponds to the input matrix, $\mathbf{W_{rec}}$ is the matrix of the recurrent connection weights. These two matrices are randomly initialized and then fixed during training. The parameters $\alpha$ is called the leak rate.

\begin{equation} \label{eq_reservoir}
    r_t \longleftarrow (1-\alpha)r_{t-1} + \alpha \tanh( \mathbf{W_{rec}}r_{t-1} + \mathbf{W_{in}}x_t) 
\end{equation}

We add a constant component to the reservoir state and then multiply this intermediate vector $s_t$ by the output matrix $\mathbf{W_{out}}$ to get the output vector $y_t$ as described in the equations \ref{eq__out_reservoir}. The output matrix $\mathbf{W_{out}}$ is optimized to get the output vector $y_t$ the closer to the teacher vector.

\begin{equation} \label{eq__out_reservoir}
    s_t \longleftarrow \begin{pmatrix} 1 \\ r_t \end{pmatrix} \hspace{10mm} y_t \longleftarrow \mathbf{W_{out}}s_t
\end{equation}

\subsubsection{Train ESN with FORCE learning} \label{force_learning}

The readout matrix was learnt using FORCE learning \cite{sussillo2009generating}. It's a second-order learning algorithm that is more biologically plausible than LSTMs because they do not use \textit{unfolding time} while training the network and they do not back-propagate errors\footnote{Simpler first order algorithms such as Least Mean Squares (LMS) are considered more biologically plausible than FORCE learning, but are much slower to converge.}. %
It works by updating the weights to keep the error as low as possible, as precised in equation (\ref{force_w_update}). 
To control the amount of change at each step for each coefficient of the matrix $\mathbf{W_{in}}$, the FORCE algorithm uses a matrix $\mathbf{P}$ that can be seen as adaptable learning rates.
This matrix is initialized along the equation (\ref{p_matrix_init}) where $\epsilon$ is a regularization coefficient. Then, $\mathbf{P}$ is updated with the equation (\ref{p_matrix_update}). Because the weights are updated after each training example, it's a form on online learning. $e_-(t)$ is the error between the prediction of the network and the ground truth at time $t$. 
\begin{equation} \label{force_w_update}
    \mathbf{W_{out}}(t) = \mathbf{W_{out}}(t-1) - e_-(t)\mathbf{P}(t)r_t
\end{equation}

\begin{equation} \label{p_matrix_init}
    \mathbf{P}(0) = \frac{\mathbf{I}}{\epsilon}
\end{equation}

\begin{equation} \label{p_matrix_update}
    \mathbf{P}(t) = \mathbf{P}(t-1) - \frac{\mathbf{P}(t-1) r_t r_t^T \mathbf{P}(t-1)}{1+r_t^T\mathbf{P}(t-1)r_t}
\end{equation}

\subsection{ESN: Final Learning and Continuous Learning}

As what was done in \cite{juven2020cross}, we used two different procedures to train the ESN.
\begin{itemize}
    \item The final learning (FL) procedure is when we apply the FORCE algorithm on the reservoir state after the \emph{last} word of the sentence (when \emph{END} is inputted. It is the usual learning procedure for ESNs trained on symbol sequences.
    \item The continuous learning (CL) procedure is used for interpretation purpose. In this case, we apply the FORCE algorithm on the reservoir state after \emph{each} word of the sentence. The ESN is constrained to predict the output even if it has access to a partial part of the sentence. This procedure worsen performances but leads to interpretable intermediate outputs of the ESN. Outputs can thus be seen as kind of probabilities of the training corpus.
\end{itemize}

\subsection{Long-Short Term Memory networks}

\begin{figure}
    \centering
    \includegraphics[width=\textwidth]{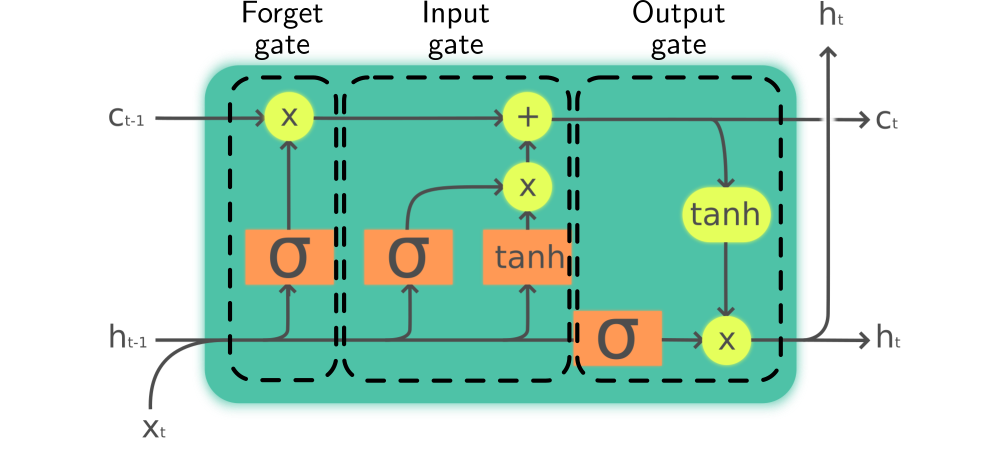}
    \caption{Architecture of one LSTM cell. A LSTM is composed of special cells $c_t$ that keeps relevant features and the hidden state vector $h_t$ that output information towards other recurrent cells and the output layer. %
    Three gates are used to decide which information to store in the cell, to forget or what to output.  }
    \label{fig:lstm_archi}
\end{figure}

LSTM networks \cite{hochreiter1997long} are a type of RNN designed to overcome the vanishing gradient problem.
They work by using gates to handle information flow between one unit own \enquote{memory} and the other units. The details of the architecture are depicted in figure \ref{fig:lstm_archi}.
For the LSTM used in this contribution, the equations ruling the architecture are: 

\begin{align} \label{lstm_eq}
    f_t &= \sigma(\mathbf{W_{f}} x_t + \mathbf{U_{f}} h_{t-1} + b_f) \\
    i_t &= \sigma(\mathbf{W_{i}} x_t + \mathbf{U_{i}} h_{t-1} + b_i) \\
    o_t &= \sigma(\mathbf{W_{o}} x_t + \mathbf{U_{o}} h_{t-1} + b_o) \\
    \tilde{c}_t &= \tanh(\mathbf{W_{c}} x_t + \mathbf{U_{c}} h_{t-1} + b_c) \\
    c_t &= f_t \circ c_{t-1} + i_t \circ \tilde{c}_t \\
    h_t &= o_t \circ \tanh(c_t)
\end{align}

$\sigma$ is the sigmoid function. $f_t$, $i_t$ and $o_t$ are respectively the forget, input and output gates. $c_t$ and $h_t$ are the inner states of the LSTM: respectively the cell vector that can be seen as the memory of the LSTM and the hidden state vector which is also the output of the LSTM that is passed to the following layers. Finally $\tilde{c}_t$ is called the cell input activation vector. 

The matrices $\mathbf{W_{f}}$,$\mathbf{W_{i}}$,$\mathbf{W_{c}}$  and $\mathbf{W_{c}}$ are used the weights of the connections from the input vector $x_t$ to, respectively, the forget, input, output gate and the activation vector. The matrices $\mathbf{U_{f}}$,$\mathbf{U_{i}}$,$\mathbf{U_{o}}$ and $\mathbf{U_{c}}$ are the weights of the recurrent connections of the LSTM, they are used to compute the values of the gates and the activation vector from the hidden state of the LSTM at the previous step $h_{t-1}$. These eight matrices along with the biases $b_f$, $b_i$, $b_o$ and $b_c$ are the parameters to be optimized by the learning procedure.

\subsection{Cross-Situational Learning}

\subsubsection{Task description} \label{task-description}
\begin{figure}
    \centering
    \includegraphics{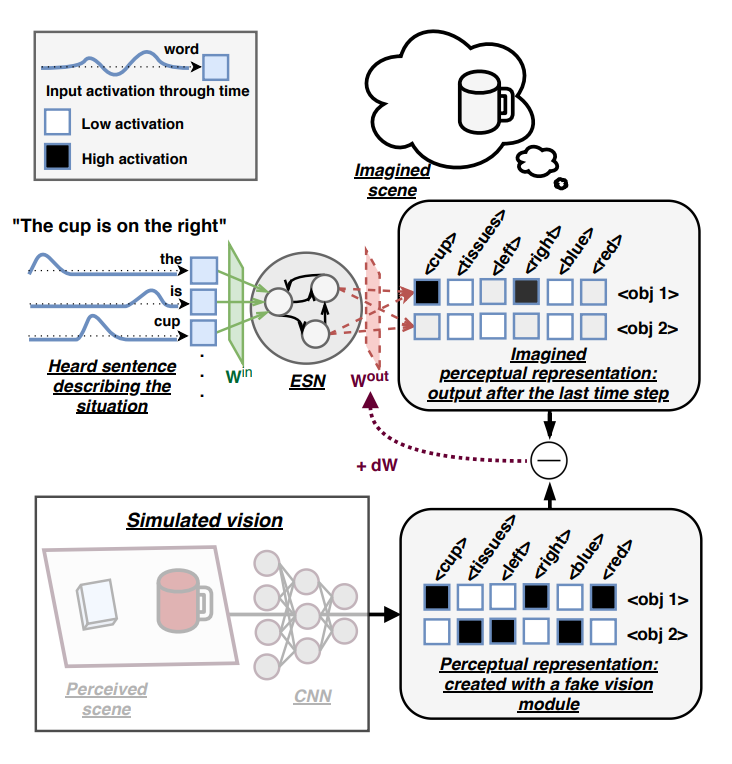}
    \caption{The Cross-Situational Learning (CSL) learning procedure for the ESN architecture.
    The model has to reconstruct an imagined scene from the sentence given word by word. The simulated vision creates a perceptual representation corresponding to the full description of objects in the scene. This representation is used as target outputs for the reservoir, even if the sentence only partially describes the objects in the scene, or if it describes only one object. This particular set-up creates cross-situational learning conditions similar to the ones children are facing. The set-up, input and target outputs were the same for the LSTM experiments.
    }
    \label{fig:csl_task_description}
\end{figure}

The task chosen for the comparison between the two models is called Cross-Situational Learning. The models are shown pair of sentences and visual perception describing objects in a scene. The goal is to learn mappings between words and the corresponding visual concepts based on their co-occurrences. In input, the sentences are represented as a sequence of words encoded using one-hot vectors. We worked on sentences with a fixed size vocabulary of size $V$. Each word in the input sentence was encoded with a one-hot $V$- dimensional vector.
The output consists in a vector representing the visual element of the scene described by the sentence. 
Each object has $Q$ possible associated visual concepts  (e.g. we used color, position and object category, $Q=3$ in this case). Each concept $c$ has $K_c$ possible concept value (e.g. \enquote{right}, \enquote{middle} and \enquote{left} for the position concept). We define a concept vector describing an object such as there is a $1$ in the corresponding concept value component if and only if the object has this concept value. The output vector is then the concatenation of the concept vectors of the first and second object, it is a $D$-dimensional vector with $D := 2\sum_{c=1}^{c=Q}{K_c}$.
An illustration of the task can be found in figure \ref{fig:csl_task_description}.
The sentence used as an input were generated using a hand made context free grammar described in figure \ref{fig:grammar}.

\begin{figure}[ht]
    \centering
    \footnotesize \begin{align*}
\nonterminal{OBJ}  \rightarrow & \terminal{ cup }  | \terminal{ bowl } | \terminal{ orange } | \terminal{ glass }\\
\nonterminal{COL}  \rightarrow & \terminal{ red } | \terminal{ orange } | \terminal{blue} | \terminal{ green } \\
\nonterminal{POS}  \rightarrow & \terminal{ left } | \terminal{ middle } | \terminal{ right }\\
\nonterminal{THE}  \rightarrow & \terminal{ a } | \terminal{ the} \\
\nonterminal{ THIS}  \rightarrow & \text{ (} \terminal{this } | \terminal{ that} \text{)}\\
\nonterminal{SENTENCE-1-OBJ} \rightarrow & \nonterminal{ THIS } \terminal{is} \nonterminal{ THE } (\nonterminal{COL})? \nonterminal{ OBJ}\\
& | \nonterminal{ THE OBJ } (\terminal{on the} \nonterminal{ POS})? \terminal{ is } \nonterminal{COL}\\
& | \nonterminal{ THE } \text{(} \nonterminal{COL} \text{)?} \nonterminal{ OBJ } \terminal{is on the} \nonterminal{ POS}\\
& | \terminal{ there is } \nonterminal{ THE } \text{(} \nonterminal{COL} \text{)?} \nonterminal{ OBJ } \terminal{on the} \nonterminal{ POS}\\
& | \terminal{ on the} \nonterminal{ POS } (\terminal{there})? \terminal{ is } \nonterminal{THE } (\nonterminal{COL})? \nonterminal{ OBJ} \\
\nonterminal{SENTENCE} \rightarrow & \nonterminal{ SENTENCE-1-OBJ} \\
& | \nonterminal{ SENTENCE-1-OBJ } \terminal{and} \nonterminal{ SENTENCE-1-OBJ}
\end{align*}
    \caption{Grammar used to generate the corpus.
    The grammar can generates sentences describing one or two objects depending on the scenario. Note that the word \enquote{orange} is used both as a noun and as an adjective: it is called a \emph{polysemous} word. The total number of different sentences that could be generated is 473344 $(= 688^2)$.}
    \label{fig:grammar}
\end{figure}

\subsubsection{Error measurement}
The teacher vectors are not a perfect representation of what the model needs to output. Indeed the visual representation can contain more information about the scene than what is described in the sentence. This is why we cannot simply quantify the performances of a model with the distance to the desired teacher vector. Thus to measure error we used two metrics: the valid and exact errors presented in \cite{juven2020cross}. 

We begin by extracting a discrete representation of the visual scene from the output vector of the model. To do so, for each concept, (e.g. the position of the second object) we use the softmax function to translate the corresponding activations (e.g. the activations of the three output neurons coding for the left, middle and right position of the second object) to probabilities. Next, we consider that if all the probabilities are below a given threshold, then it means that this concept was not present in the input sentence. If at least one probability is above the threshold, we consider that the concept value is the one corresponding to the maximal probability.

\begin{figure}
    \centering
    \includegraphics[scale = 0.4]{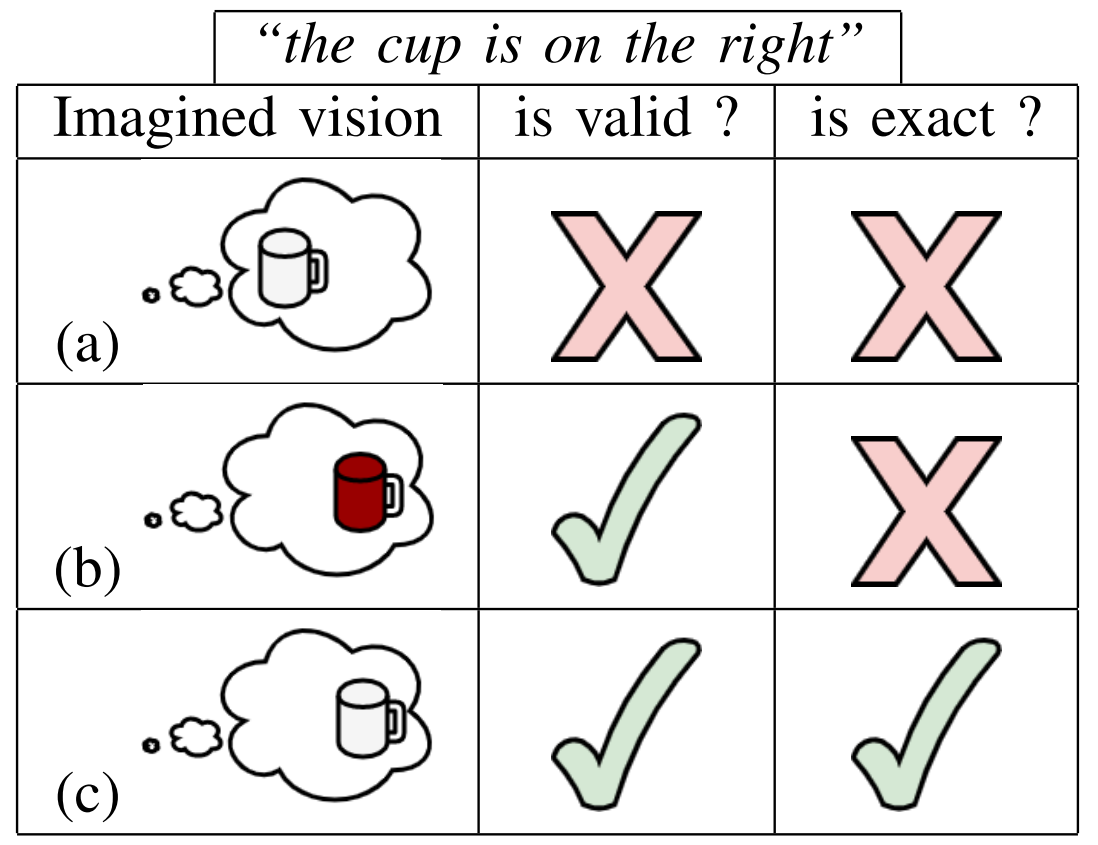}
    \caption{Evaluation examples of different scene representations. The imagined scene (a) is not valid nor exact because the cup is not on the right. (b) is not exact because the sentence does not precise the cup color but the scene precises it to be red.}
    \label{fig:different_errors}
\end{figure}

For a concept $c$, the threshold is fixed to $1.3/K_{c}$ throughout the paper. It enables a fair comparison between the two models. The choice of this value can be seen as if it was part of the task specification. For more details, we investigate the influence of this choice in supplementary material subsection \ref{subsec-threshold_factor_influence}.
After having extracted this discrete representation, we can compare it with the ground-truth information contained in the input sentence. At this step there are three possibilities, depicted in figure \ref{fig:different_errors}:
\begin{enumerate}[(a)]
    \item The model's representation lacks some information contained in the sentence. The output is then \emph{not valid}.
    \item The model's representation has all the information but also contains concepts values that were not in the sentence. It's qualified as \emph{valid}.
    \item The representation contain all the information of the sentence and nothing more. It's then \emph{exact}.
\end{enumerate}

\subsubsection{Theoretical model} \label{theoretical_model}

To compare the performances of the machine learning models (ESN and LSTM) with a more naive one, we constructed a theoretical model to provide a reference point in performance comparison.
It works by memorizing associations between unordered sets of keywords 
and teacher outputs, one set for each object. For instance, the sentence \enquote{BEGIN there is a red cup and this is a glass on the left END} will be split in two keyword sets: ${red, cup}$ and ${glas, left}$.
During training, this model constructs a table of associations between keywords sets and corresponding half teacher vector: because teacher vectors encode the characteristic of both objects, the keyword sets are associated to the corresponding part of the teacher vector. In other words, the set of keywords related to the first object is associated to the first half of the teacher output, and the set of keywords related to the second object is associated to the second half of the teacher output.
Then, during testing, every sentence is split along the \enquote{and} word and the two parts are processed independently. If the two keywords sets are present in the association table, the model output the concatenation of the two corresponding half teacher vectors. Otherwise if a keyword set is not present, it outputs an empty output vector.

\subsection{Details of the training corpus}

For both models, the training and test sets were composed of 1000 sentences: 700 sentences with two objects and 300 with one object. 

\subsubsection{The ESN implementation}

We used the ReservoirPy toolbox\footnote{ReservoirPy repository: \url{https://github.com/neuronalX/reservoirpy}}, for the implementation of ESNs \cite{trouvain2020reservoirpy}. The hyper-parameters were optimized through random search by Juven \& Hinaut in \cite{juven2020cross} for one epoch of learning. Their values can be found in table \ref{tab:hp_table_esn}. We used ESNs with 1000 recurrent units. ESNs were trained using the FORCE learning algorithm detailed in subsection \ref{force_learning}.

\begin{table}
 
  \centering
  \begin{tabular}{lll}
    \toprule
    Hyper-parameter     & Value \\
    \midrule
    Spectral Radius            & 1.1    \\ 
    Leak Rate                  & 0.05   \\ 
    Sparsity                   & 0.85   \\ 
    Regularization coefficient & 3.2e-4 \\ 
    Input Scaling              & 1.0    \\ 
    \bottomrule
  \end{tabular}
  \vspace{2mm}
  \caption{Hyper-parameters used for the Echo State Network, optimized through random search in \cite{juven2020cross}}
  \label{tab:hp_table_esn}
\end{table}

\subsubsection{The LSTM implementation}

 We used the Keras implementation of LSTM and the Adam optimizer \cite{kingma2014adam} with default parameters. The standard LSTM used was composed of 20 units. In the Keras implementation, the sigmoid function was replaced by an affine approximation called the hard sigmoid function in order to optimize performances. We trained it during 35 epochs composed by training sentences organized in batches of size 2. The number of epochs for training was chosen with early stopping method: we conducted experiment to determine when the error on the validation set was at its lowest. 
 The use of dropout did not help with the convergence during learning for the 4-object dataset so we did not use it for this task. 
 These choices were made to create a LSTM that performs well on the 4-objects dataset. After the output of the LSTM we used a feed-forward layer to the output layer: for a LSTM of dimension $N$, we compute the output vector $y_t = \mathbf{W_y} h_t$ were $\mathbf{W_y}$ is a $[D\times N]$ parameter matrix. $D$ is the output dimension, defined in subsection \ref{task-description}. %
 
 As for the ESN, we did not use any specific activation function (i.e. linear activation $f(x)=x$).

\section{Performance Results}
\label{sec-perf-results}

\subsection{4-objects dataset}

We run the experiment with the setup described above on a laptop architecture. The sentences were generated by the grammar described in figure \ref{fig:grammar}. The models ran on an Intel Core i5-8265U CPU running at 1.6GHz. The results of 30 independent runs (corresponding to various randomly initialized instances) of the models can be found in table \ref{tab:table_perf}. We found that both models are able to successfully learn the task with low error. However the LSTM outperforms the ESN on both valid and exact error, even if errors are very low for both models (LSTM: $0.037\% \pm 0.10$; ESN: $0.21\% \pm 0.15$). But this performance gain comes with an average training time which is nine times bigger. This computation difference is present despite the fact that the LSTM architecture used here was pretty small with only 3,742 trainable parameters (to be compared with the 22,000 parameters -- 1000 recurrent units x 22 outputs -- to be learnt in the readout matrix of the ESN). The Back-Propagation Trough Time (BPTT) 
algorithm used to optimize the LSTM needs to account for all the time steps in the input sequence to update the parameters. Due to this costly operation, the optimization of the small number of parameters in the LSTM took more time than the FORCE learning on a larger number of parameters. To conclude, the ESN appears as a computationally efficient model for this task while the LSTM can achieve slightly better performances at the expense of more expensive computations. 

\begin{table}
 
  \centering
  \begin{tabular}{l cc cc cc}
    \toprule

    ~ & \multicolumn{2}{c}{Valid error} & \multicolumn{2}{c}{Exact error} & \multicolumn{2}{c}{Training time (s)} \\
    \cmidrule(lr){2-3} \cmidrule(lr){4-5} \cmidrule(lr){6-7}
    
    Model & Mean Value & Std Dev. & Mean Value & Std Dev. & Mean Value & Std Dev. \\
    
    \midrule
    LSTM    & \textbf{0.037}\% &  0.10  & \textbf{1.5}\% & 1.1 & 410 & 99  \\
    ESN (Final Learning)     & 0.21\% &  0.15  & 5.7\% & 1.0 & \textbf{47.1} & 1.3  \\
    ESN (Continuous Learning)     & 2.4 \% &  0.60  & 10.6\% & 1.6 & 379 & 57 \\
    \bottomrule
  \end{tabular}
  \vspace{2mm}
  \caption{Performance of the ESN and LSTM on the CSL task. The results are averaged on 30 different instances of the models. The final learning method is the most cost-efficient training method for learning these type of task whereas the continuous learning was used for visual interpretation purposes.}
  \label{tab:table_perf}
\end{table}

\subsection{Performance with more objects in the vocabulary}

\begin{figure}
    \centering
    \includegraphics[width = \textwidth]{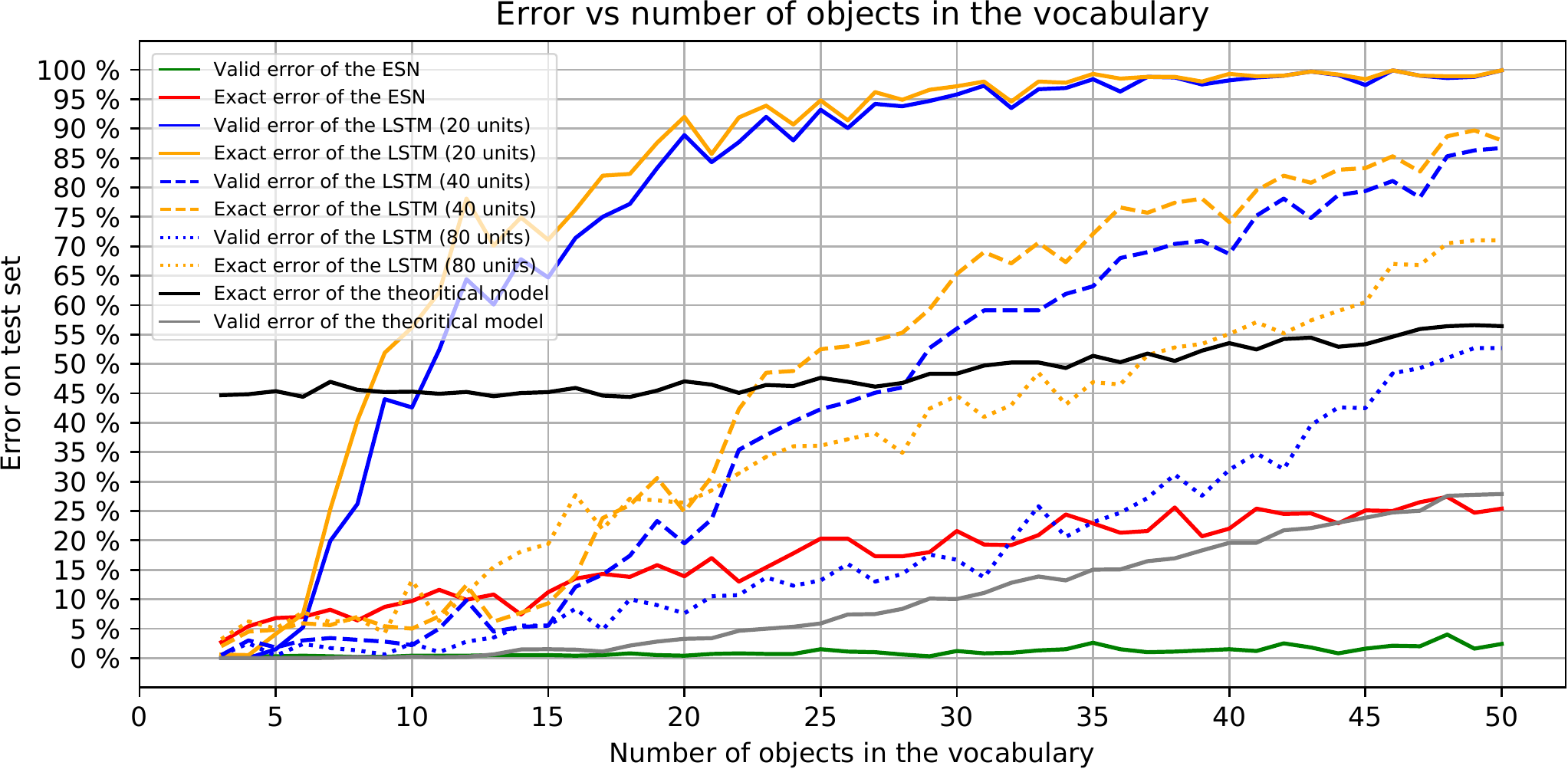}
    \caption{Comparison of the performance of 5 models (1 ESN + 3 LSTMs + theoretical) for different number of objects in the dataset. We can see that the small LSTM (20 units), optimized to perform well on a dataset with 4 objects, is not able to keep good performance with a higher number of objects. The medium LSTM (40 units) trained for longer with dropout is able to outperform the ESN until 15 objects. The bigger LSTM (80 units) limits the rise of the error compared to the other LSTM. However, it comes with poorer performances even for a small number of objects. The ESN is able to keep an error below the theoretical model and all the LSTMs despite the fact that its hyper-parameters were optimized for the 4-object dataset.}
    \label{fig:effect_nb_obj}
\end{figure}

To compare the abilities of the models to be trained on datasets with more features, we run experiments by training the models on different datasets with a variable number of objects in their vocabulary. The results of this experiment are shown in figure \ref{fig:effect_nb_obj}. If we used the LSTM architecture optimized %
for the 4-objects dataset, 
we can see that the error explodes as soon as we increase a little bit the vocabulary size (i.e. number of objects) compared to the 4-object dataset for which it was designed. This drop in performance happens much more quickly than for the ESN that succeeds at keeping its error below the one of the theoretical model (presented in subsection \ref{theoretical_model}). The ESN learnt quicker than the theoretical model because it learnt to decorrelate the concepts whereas the theoretical model only remembers fixed combinations. Moreover, because the theoretical model remembers teacher outputs without any further processing, it often predicts more information than what is precised in the sentence (i.e. high exact error): the theoretical model has no sense of probabilities. This is why it always has high exact error even on small datasets.

We then conducted another experiment with a 40-unit LSTM, trained with a dropout of 0.2 on 50 epochs. 
By comparison on a validation set, we found that the dropout value of 0.2 was enough to avoid over-fitting for this number of epoch\footnote{The value 0.2 is a common value used to train LSTMs in the literature.}.
We found that this bigger model was able to outperform the ESN on exact error until 15-objects datasets. After that, even the exact error began to rise higher than the theoretical model. 
Another LSTM model with 80 units was tested. We found that it only needed to be trained during 15 epochs to get optimal performances, before over-fitting on the 4-object dataset. The use of dropout did not seem to help here. We found that this model globally keeps the error lower than the two other LSTMs, especially for high number of objects. Nonetheless, the ESN also outperformed it for all the range tested.

In the end, the ESN is able to keep the error low on challenging datasets despite having hyper-parameters optimized to perform well on a 4-object dataset. Whereas for the LSTMs to successfully learn these more featured datasets, we need to scale up the architecture. 

\section{Comparison of single cell activation}
\label{sec-comp-single-cell}

\begin{figure}

    \centering

    \begin{subfigure}{\textwidth}
        \centering
         \includegraphics[width=0.69\textwidth]{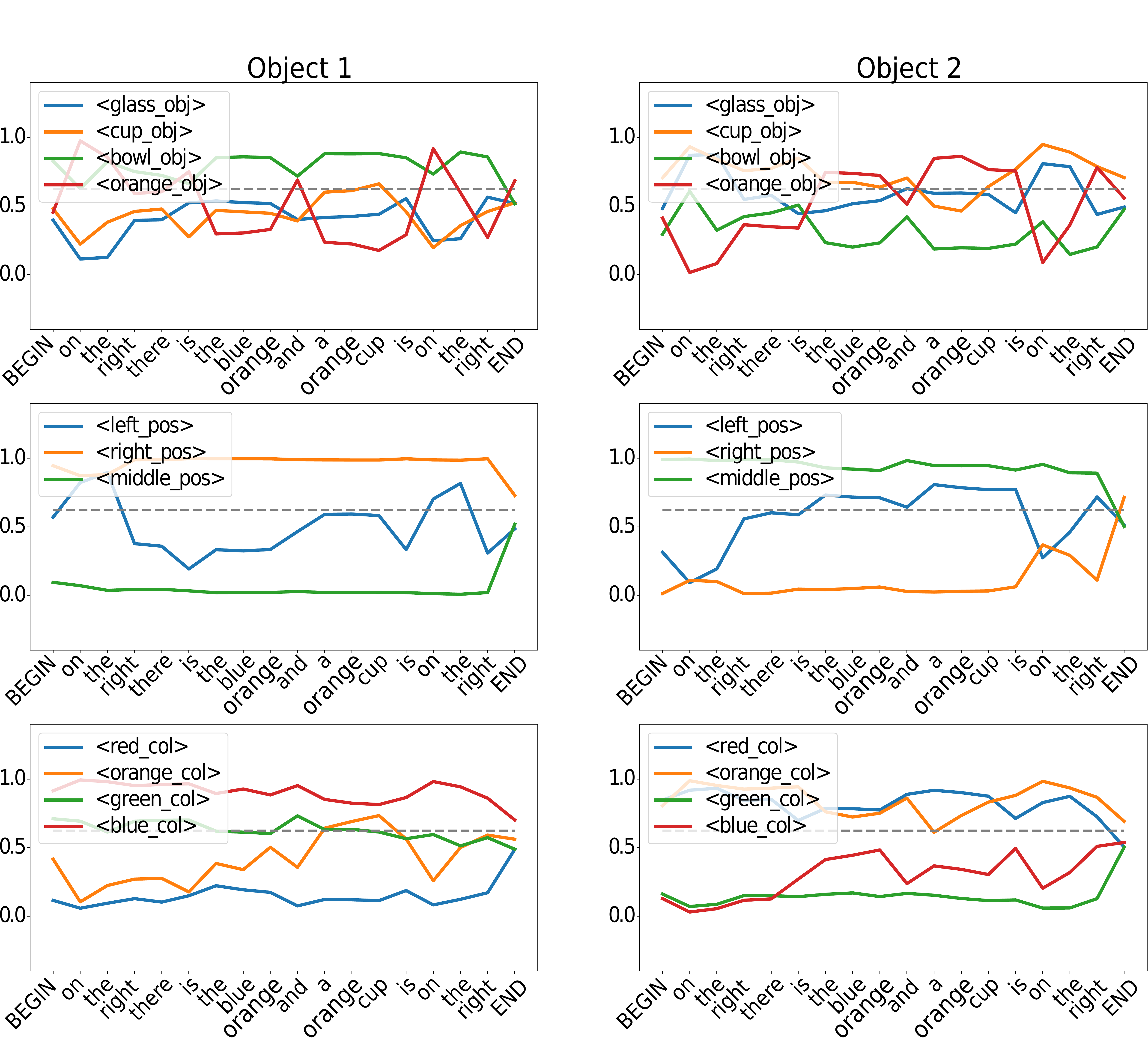}
         \caption{ The ESN outputs trained with final learning (FL). The activation are here shown after being transformed by the sigmoid function $\sigma : x \mapsto \frac1{1 + {\rm e}^{-x}}$ 
         to get bounded values. Indeed, because no constrain is imposed on intermediate outputs, they take arbitrary big or small values, far outside the range [0,1]. For this reason they are difficult to interpret.}
    \end{subfigure}

    \begin{subfigure}{\textwidth}
        \centering
         \includegraphics[width=0.69\textwidth]{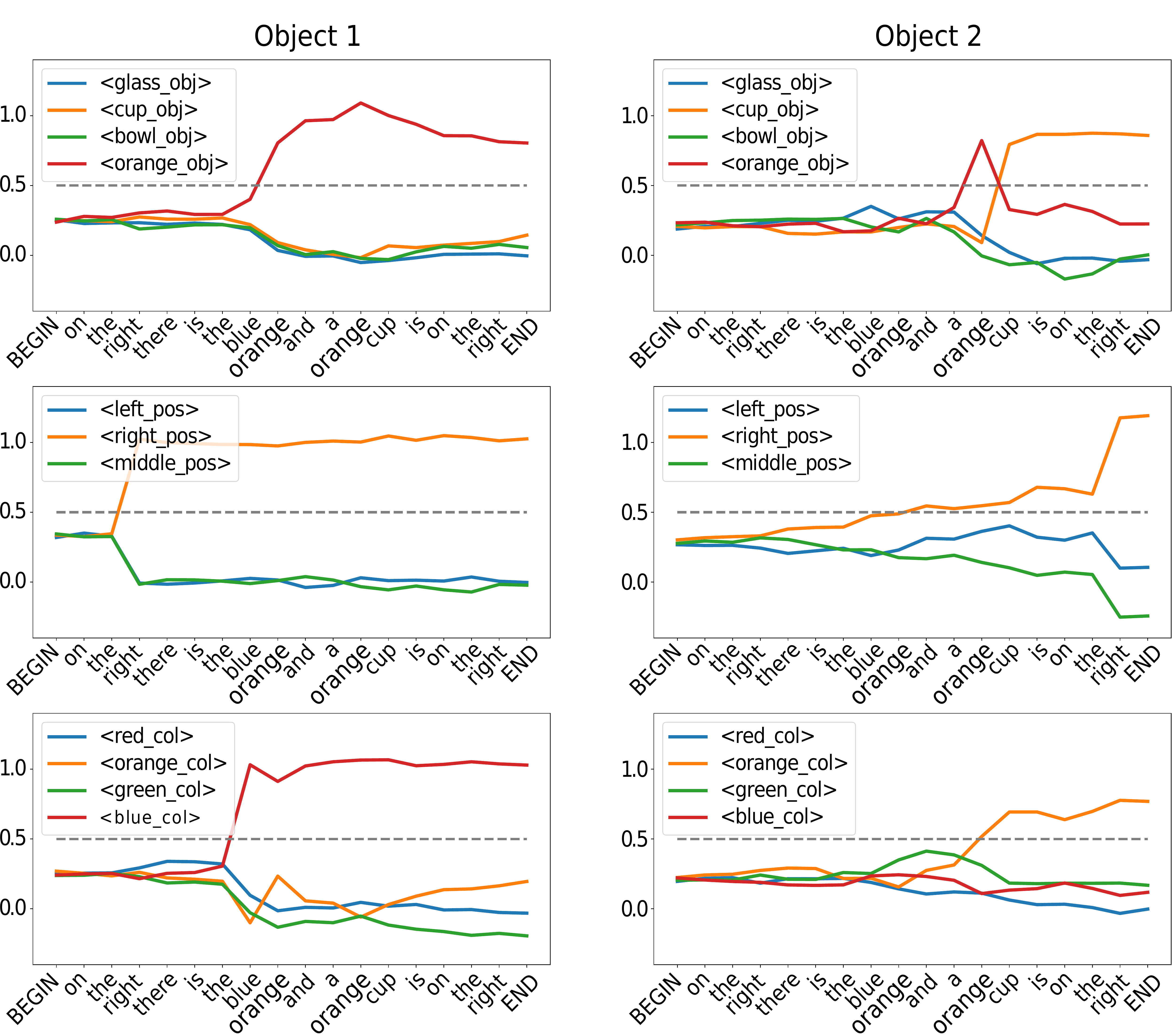}
         \caption{The ESN outputs trained with continuous learning (CL). After each word the model tries to predict the correct output. That's why we can see a jump in the correct characteristic after the related keyword is seen.}
    \end{subfigure}
\end{figure}
\begin{figure} \ContinuedFloat
    \begin{subfigure}{\textwidth}
        \centering
         \includegraphics[width=0.7\textwidth]{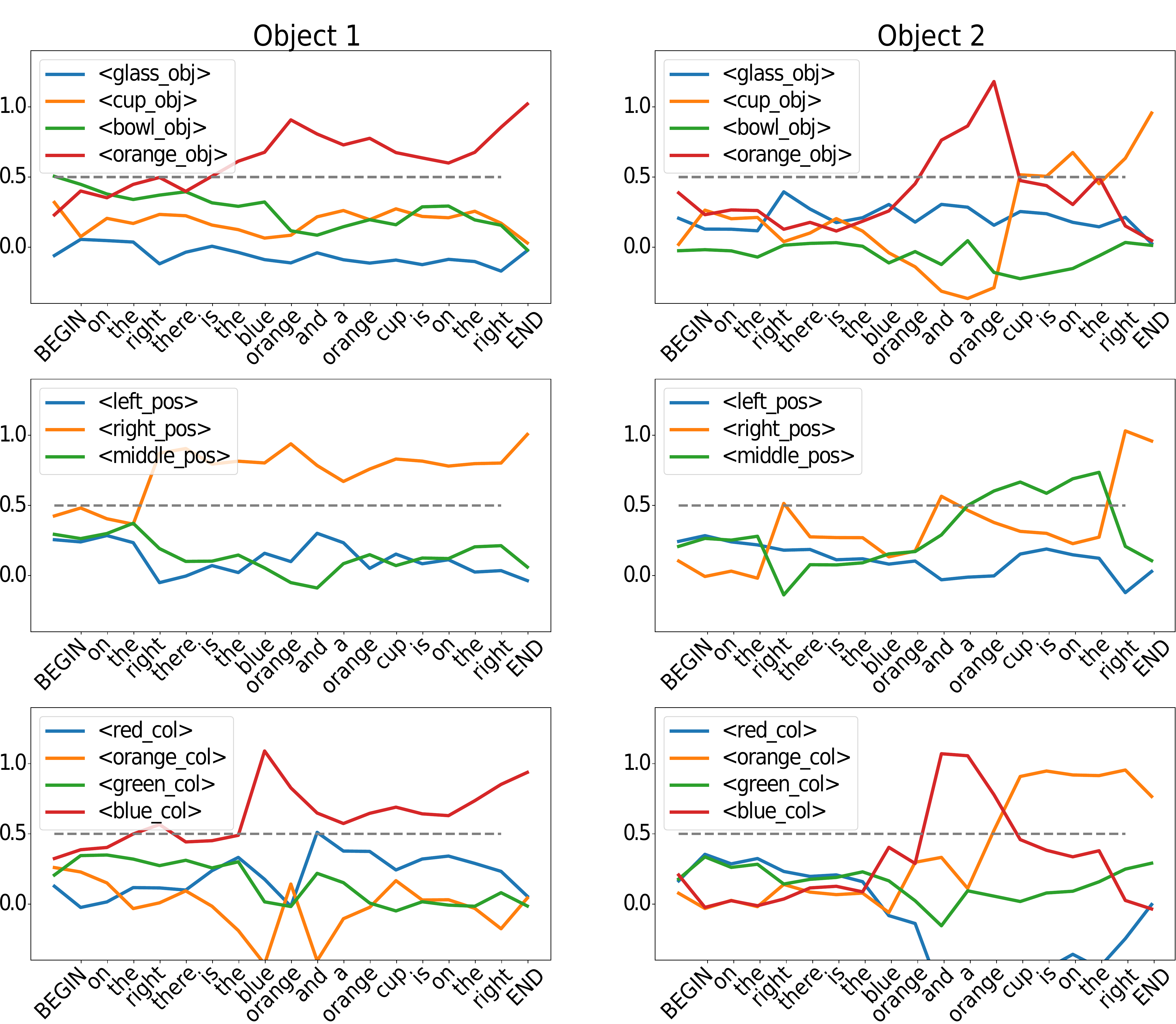}
         \caption{ Output activation of the LSTM. Even if the learning procedure is only applied at the end, because of its learning algorithm, intermediate states are also optimized. This is why we can also interpret these transitional steps: they behave similarly to the ESN trained with CL. }
    \end{subfigure}
    
    \caption{
    Evolution of the output activation during the processing of the sentence: \enquote{On the right is the blue orange and \textit{a} orange cup is on the righ}. For reference purpose, the dotted line represent an output activation of 0.5.}
    \label{fig:acti_plots}
\end{figure}

\subsection{Qualitative analysis of output units activation} 
\label{comparing_output_activation}

To get a first view of the inner working of both models, we can start by plotting the evolution of the output activation during the processing of a sentence. However, for the ESN, the intermediate output activation cannot be interpreted with the default Final Learning (FL). As we can see in figure \ref{fig:acti_plots}(a), the fluctuations seem unpredictable until the last word \enquote{END} is seen. For a correctly predicted output, the activation often \enquote{jumps} to the correct value when the last item \enquote{END} is inputed. This is due to the fact that we only apply the learning procedure at the final state, so there is no constrain on intermediate outputs.

Interestingly, this was not the case for the LSTM: the activation of a given output generally jumps as soon as the corresponding keyword is seen by the model.
It's quite unclear why the LSTM did not show similar output activation pattern as the ESN. Since there is no constrains on the intermediate output during the learning procedure, the LSTM could theoretically output the correct prediction only after the word \enquote{END}. Instead, in the figure \ref{fig:acti_plots}(c) the word \enquote{END} doesn't seem to affect significantly the output of the network. We have not yet found a clear explanation for this behaviour but we will analyze this phenomenon in more depth in subsection \ref{lstm-rssviz} thanks to our proposed RSSviz visualisation.

\label{description_effet_preidction_LSTM_imparfaite} However, because no constrains are specified on these intermediate states, these \emph{jumps} do not follow a perfect rule. For instance in the figure \ref{fig:acti_plots}(c), %
we can observe a spike in the activity of the concept <blue\_color2> (i.e. the concept activated when the object 2 is blue) when the word \enquote{blue} corresponding to the first object is seen. 

In order to interpret the transitional outputs of the ESN (i.e. during the processing of a sentence) we applied the learning procedure at each time step, similarly as in \cite{hinaut2013real, hinaut2015recurrent} but with online learning (offline learning was used in these previous studies). We call this process Continuous Learning (CL) as opposed to to Final Learning (FL).
This new learning method significantly affected the performances both in training time and in errors as shown in the table \ref{tab:table_perf}. Indeed we added constrains to the learning and we performed more computationally expensive tasks. Nonetheless, the performances are still good enough to consider that the task was successfully learnt by the model. We will use this learning method in the following analysis.

Figure \ref{fig:acti_plots} gives a qualitative insight on the way the models work. Plots b) and c) show which is the most plausible meaning of the sentence given the words already seen. For example, we can clearly see that when the word \enquote{orange} is seen for the second time, the model has not yet the information that the word will be used as an adjective. So, when this happens, for both the LSTM and the ESN trained with CL, we can see a rise (i.e. a spike) in the activation of the <orange\_object2> concept (i.e. the output neuron that should be activated when the second object is an orange). %
This spike is quickly inhibited when the following word \enquote{cup} is received: it's now clear that \enquote{orange} was in fact an adjective and not a noun. This phenomena gives us a first hint on how both models are able to deal with polysemous meaning. They both seems to consider the \enquote{orange} word as an noun when no full context is provided.

\begin{figure}
    \begin{subfigure}{\textwidth}
        \centering
         \includegraphics[width=0.8\textwidth]{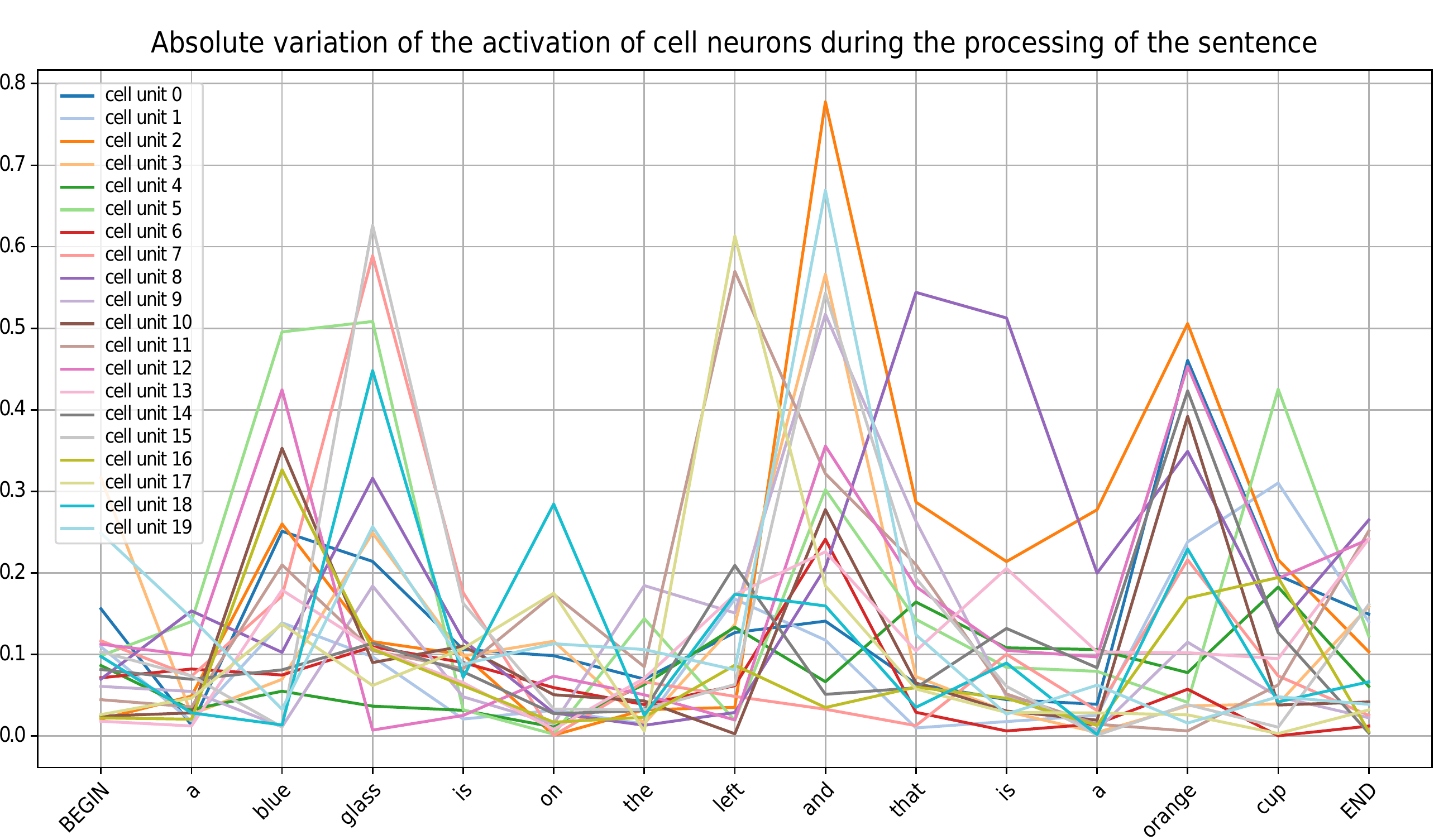}
         \caption{ Absolute variation in activity of the 20 LSTM cells. The semantic words cause significantly more variation than non semantic words. We compute the mean of the sum of the cell absolute variation on 1000 sentences\footnote{ The value are shared in the form $mean\pm standard~deviation$.}: for semantic words: 3.15\textpm0.56, for function words (except \enquote{and}): 1.50\textpm0.40. The word \enquote{and} is excluded from function words statistics because it has a special important semantic \enquote{role} for the task.
         }
    \end{subfigure}
    \begin{subfigure}{\textwidth}
        \centering
         \includegraphics[width=0.8\textwidth]{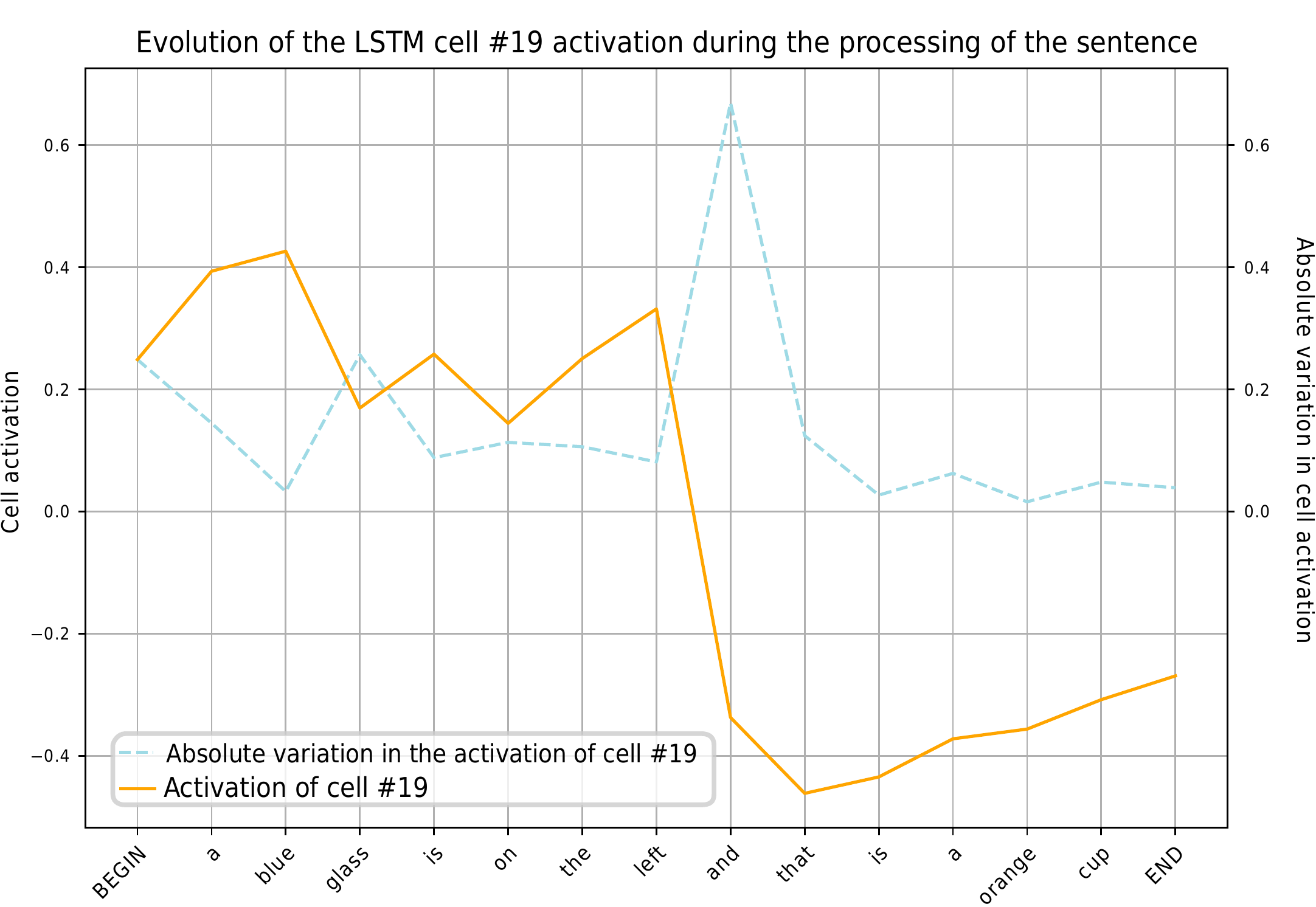}
         \caption{A LSTM cell sensitive to the delimitation of the two clauses in a 2-object sentence.
        When the word \enquote{and} is seen, its activity significantly decreases: on 1000 sentences, the word \enquote{and} causes an absolute variation of 0.70\textpm0.09, %
        more than any other word. The second most influencing word was \enquote{glass}, causing a variation of 0.30\textpm0.19, and for an average word: 0.15\textpm0.2.
        }
    \end{subfigure}

    \caption{We can find interpretable behaviours in the activation of the LSTM cells. }
    
    \label{fig:cells_LSTM}
\end{figure}

\subsection{Understand the activation of hidden units} \label{sec-understand_hidden_cells}
\subsubsection{Interpretable LSTM cells}

We found that the LSTM has cells that keep track of the sentence clause (i.e. 1st or 2nd part of a sentence). They are able to memorize if the word seen is related to the first or to the second object. As shown in figure \ref{fig:cells_LSTM}, we observe a significant variation at the word \enquote{and}. This word is the best indicator to delimit the two clauses. %
It makes sense that this feature is important to be accurate but it's striking that this sentence structure emerges clearly in the LSTM memory. Nonetheless, this cell was not the only one to respond to the word \enquote{and}, as visible in the figure \ref{fig:cells_LSTM}.A. The special thing about this cell unit was its \emph{specific} activity to the word \enquote{and}. However, this cell alone is probably not the only one to account for the sentence structure.

With a more general view, because the LSTM learnt an inner representation of the sentence optimized for the task, the variation in its overall cell activity is not constant for all words. We can see in figure \ref{fig:cells_LSTM} that words with semantic meaning, i.e. keywords like \enquote{right}, \enquote{glass} or \enquote{orange} cause more variation than function words like \enquote{that}, \enquote{there}, \enquote{is}. This makes sense because they are the most useful words to predict the outputs.

\subsection{Top 5 reservoir units are like trained LSTM cells} \label{analysis_of_reservoir_units_activation}

\begin{figure}
    \begin{subfigure}{\textwidth}
        \centering
         \includegraphics[width=0.8\textwidth]{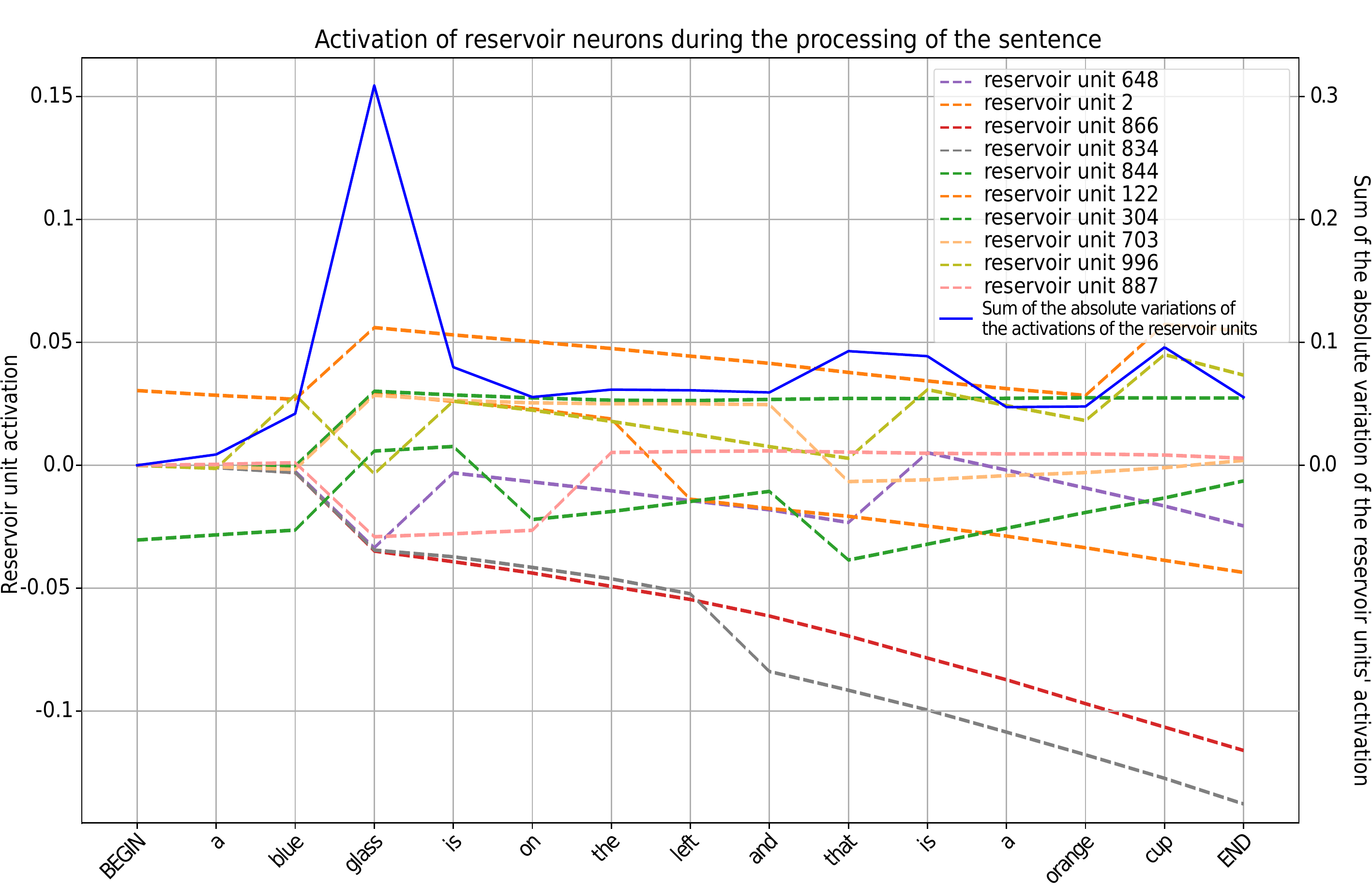}
         \caption{
         \textit{Dashed line}: activity of the 10 reservoir units the most connected to the output units <glass\_obj1>. 
        \textit{Full line}: summed absolute variation in activity of these units. 
        These reservoir units react to the word \enquote{glass} more than any other: on 1000 sentences, the summed absolute variation for the word \enquote{glass} was on average 0.312\textpm0.004\footnote{ The value are shared in the form $mean\pm standard~deviation$.} and the second most affecting word was \enquote{green}, causing a mean variation of 0.097\textpm0.014.}
    \end{subfigure}
        \begin{subfigure}{\textwidth}
        \centering
         \includegraphics[width=0.8\textwidth]{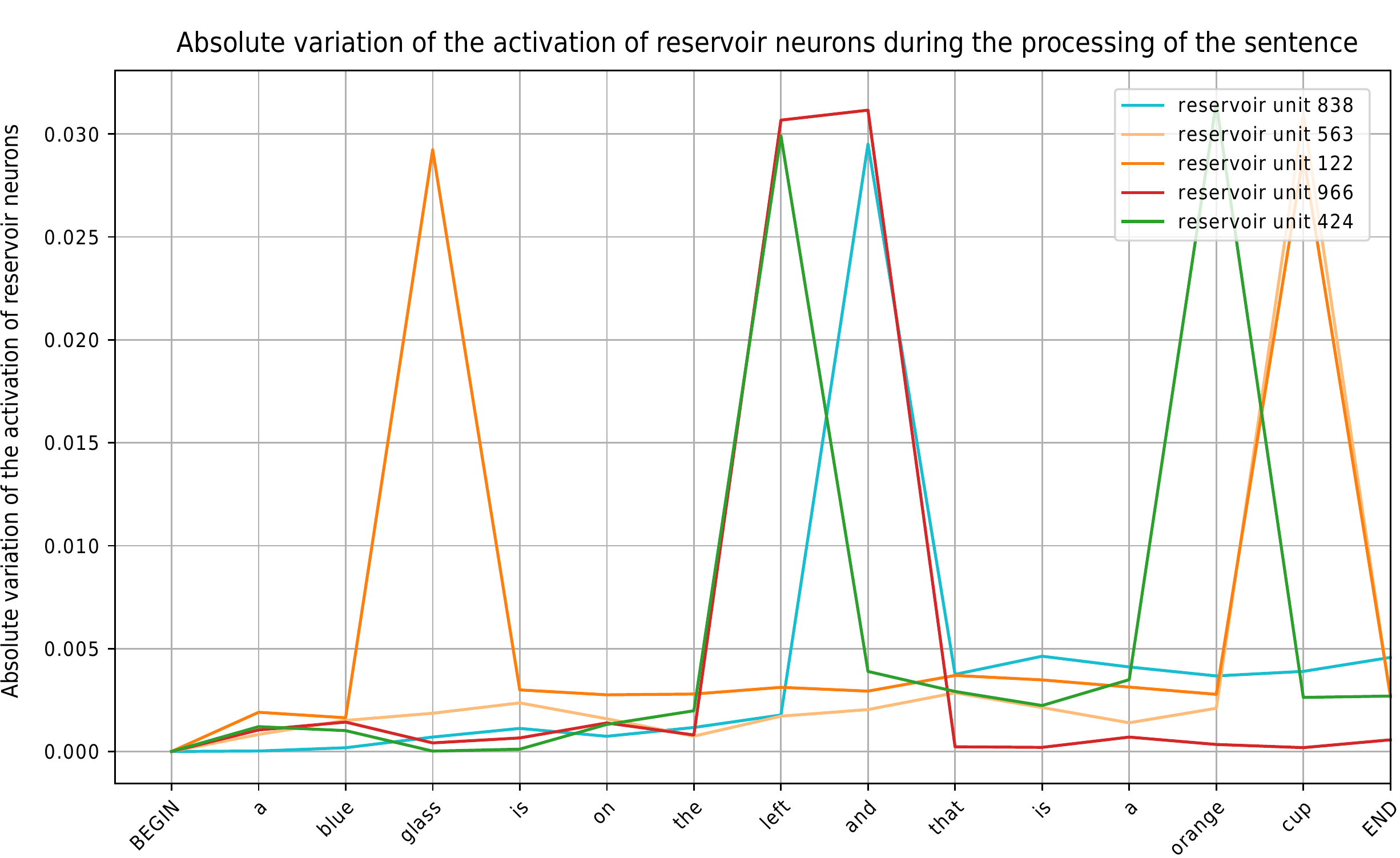}
         \caption{The 5 multipurpose reservoir units were, like the LSTM cells more sensitive to semantic words than non semantic words: their summed absolute variation for semantic word was 0.0306\textpm0.022 and only 0.0096\textpm0.005 for non semantic words (the variation is here more important than for the LSTM cell because, according to the graph, these reservoir units seem to be specialized for a few semantic words. There is some semantic words such as the word \enquote{blue} in this sentence, to which they do not react.). %
        Moreover the reservoir unit 838 was sensitive to the word \enquote{and} more than any other: \enquote{and} causes on average an absolute variation of 0.030\textpm0.0003, and for the second most sensitive word, \enquote{left}: 0.0031\textpm0.0016. }
    \end{subfigure}
    \centering
    \caption{
    Activations of reservoir units with the biggest weights to the output layer. 
    }
    \label{fig:specialized_reservoir_units}
\end{figure}

Of course, we cannot apply the kind of analysis we did on the LSTM directly to the ESN. We do not use any feedback in our reservoir so the states of the reservoir are fully determined by its random weights and the inputs received. In fact, the learning process happens by combining the useful activities given the random projections of the inputs done in the reservoir. 
To make sense of this selective process, we used a method akin to feature selection. For a given output unit, we searched for the reservoir units from which is has the strongest weights. This can help understand on which reservoir (sub-)activation pattern this particular neuron \enquote{focuses}.
We used such procedure for the \emph{<glass\_object\_1>} neuron, the results are shown in Figure \ref{fig:specialized_reservoir_units} A. We can see that the majority of these reservoir units are at least partially interpretable: they are highly correlated with the appearance of the keyword corresponding to the object characteristic.

Afterwords, to get a more general view, we searched for the reservoir units that were the most multipurpose (a concept which could be related to the concept of \emph{mixed-selectivity} \cite{rigotti2013importance, enel2016reservoir}). More practically, we began by computing the 20 reservoir units the most connected to each output neuron. Then, we looked for the reservoir units that appeared most in the top 20. We found that 5 reservoir units appeared in the top connected units for more than 5 outputs neurons. As visible in the figure \ref{fig:specialized_reservoir_units}.B, these units seem to be sensitive to semantic words, much like the trained LSTM cells. Moreover if we look at the reservoir unit 838, we can see that it reacts specifically to the word \enquote{and}, like the LSTM cell identified in figure \ref{fig:cells_LSTM}.

It's striking to see that even if ESNs and LSTMs are widely different architectures, to some extent, they both seems to converge toward a common strategy to tackle this task. By means of feature selection for ESNs and feature creation for LSTMs.
Indeed, ESNs create random features (with the input and recurrent layers) and then select some of them (with the output learning), while in LSTMs the creation of features is lead by the optimisation of weights (and thus internal states).

\section{Visualizing the latent space of LSTM and ESN in a low dimensional space}
\label{sec-RSS_ESN}

\subsection{Visualisation protocol}

In the previous section, the interpretation of individual neurons gave us valuable insights for both models. It reveals similar emergent mechanisms to process sentence structure. However, these types of analysis don't grasp the whole inner mechanisms of the models: we could not reduce the high dimensional information processing to a few units. That's why we attempt to create a visualisation that accounts for the whole recurrent state. We used the dimension reduction technique UMAP to create 2D maps that show the internal recurrent states with color context from the inputs or from the outputs produced. Similar techniques have already been used in diverse contexts. However, in order to easily discuss these methods, we refer to them as Recurrent States Space Visualizations (RSSviz).

\begin{figure}
    \centering
    \includegraphics[width = \textwidth]{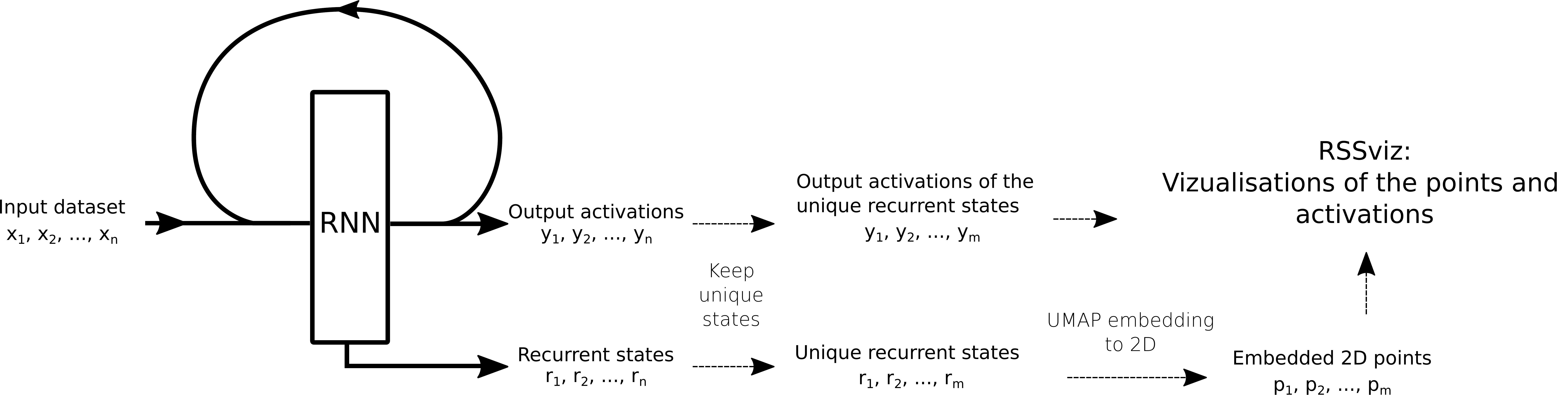}
    \caption{Protocol to create RSSviz. We used only unique recurrent states to avoid a resultant space distorted toward states that appears with high frequency.
    } %
    \label{fig:RSS_schema}
\end{figure}

As illustrated in figure \ref{fig:RSS_schema}, we gathered the recurrent states of a RNN to create a low dimensional representation thanks to UMAP \cite{mcinnes2018umap}. Only unique recurrent states were sampled in order to avoid distortion in the resultant space. Indeed, some states were much more present than other. For example, because sentences often begin in the same way, the states corresponding to partial sentences like \enquote{BEGIN the} or \enquote{BEGIN there is} were much more present than those corresponding to the final state of a sentence. If we kept all the recurrent states for the calibration of UMAP, we would have a space that give wide surfaces to these states that carry little information.

Then, we created different representations using color on these points. For example, we used color scale to represent output activation or position in the sentence. In the following figures, if not precised, The UMAP embedding were made on the unique recurrent states gathered on 1000 random sentences (300 with one object, 700 with two objects, the same composition as the training and test sets of the models).

We choose UMAP instead of other dimensional reduction methods like t-SNE \cite{maaten2008visualizing} or PCA for mainly two reasons. First, because it has already been successfully applied to high dimensional representations of neural networks in \cite{carter2019activation}. Second, because of its inherent qualities as computation efficiency and its preservation of the structure of the initial dataset. For all the visualisations, we used the following parameters that gave interpretable results: $n\_neighbors = 15$ and $min\_dist = 0.1$. Other combination for $n\_neighbors$ and $min\_dist$ were tested without affecting the conclusions of our study. We kept the parameters fixed for the experiments presented here to be able to compare the shape of the figures.

In fact, this protocol can be applied for virtually any artificial neural network. Indeed, as discussed in the introduction, this has already been extensively investigated in \cite{carter2019activation} for Convolutionnal Neural Networks (CNN). However, little has been done with this visualisation for LSTMs and even less, if not nothing, for ESNs. Nonetheless, we think that the range of this tool can be much wider than what it has already been used for as of now.

In the following parts of the study, we will see how the structure of this low dimensional space and the visualisation of the corresponding output activation can give us valuable information about the functioning of these two neural network models on the CSL task.

\subsection{Recurrent States Space Visualisation of an ESN}

The inner states of an ESN are fully determined by its random initialisation and the input sequence. Thus, RSSviz represents both statistical properties of the dataset and intrinsic properties of the reservoir. However, the position of the points doesn't depend on what the ESN has learnt. Instead, the ESN's RSSviz gives us a view on which state subspaces the ESN has to partition (with hyperplanes) to learn the task (i.e. to provide the correct outputs). 

\begin{figure}
    \centering
    \includegraphics[width = \textwidth]{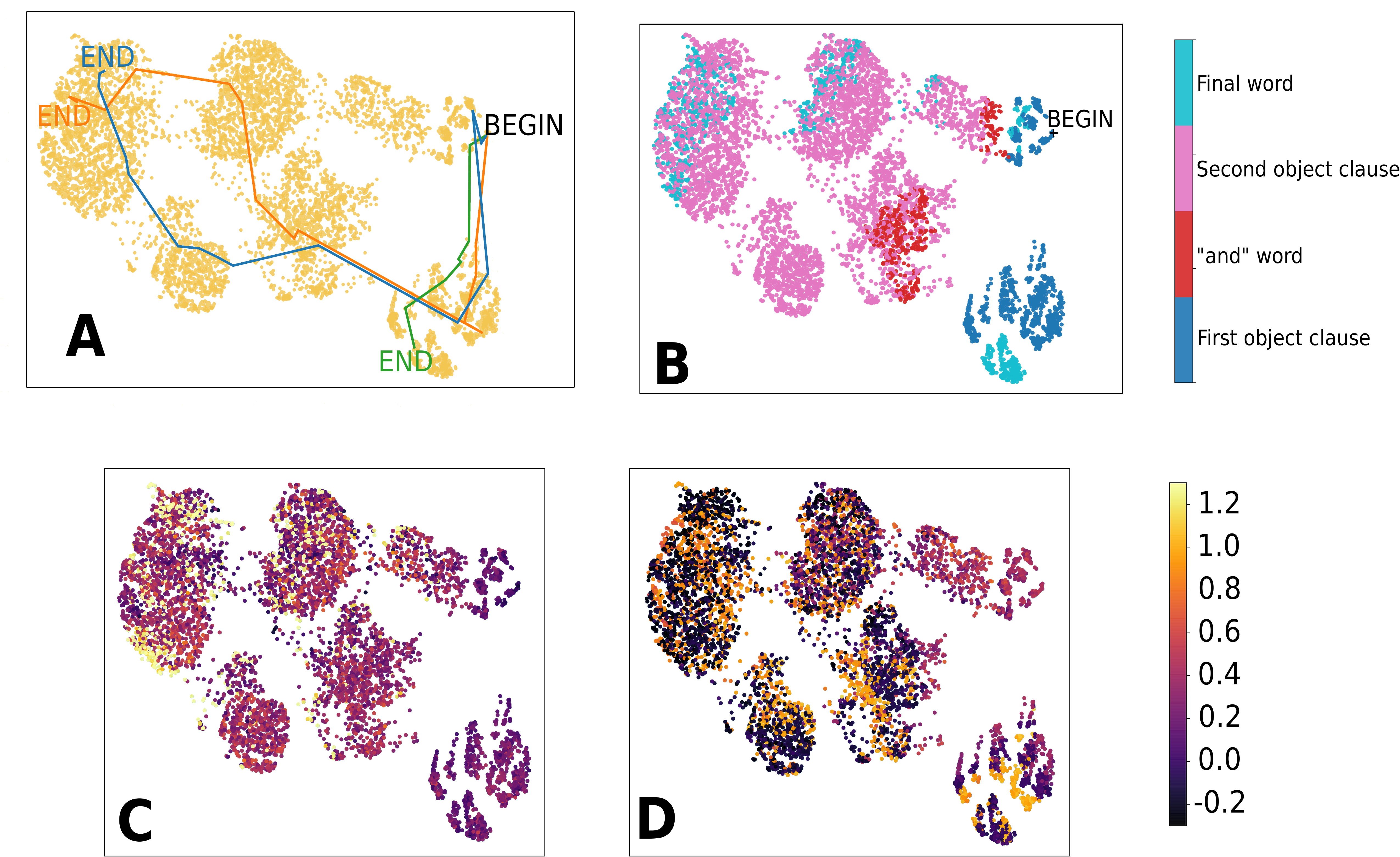}
    \caption{RSSviz for the ESN. The UMAP embedding were made on the unique reservoir states gathered on 1000 random sentences (300 with one object, 700 with two objects). 
    \textbf{A.} Each point correspond to a reservoir state. Every sentence correspond to a trajectory in RSSviz. Here 3 sentences are shown: two containing the description of two objects (blue and orange lines) and one, containing one object (green line). 
    \textbf{B.} Visualisation of the position in the sentence. We can see a clear division between the states corresponding to words in the first clause of the sentence on the right, in dark blue (the first object continent) and the pink part corresponding to the second clause (the second object continent). The red dots, corresponding to the word \enquote{and} create a border between the two continents. The light blue dots are the final states, they are often located at the extremity of the structure. 
    \textbf{C.} Activation of the concept \enquote{The 2nd object is a bowl} (<bowl object 2>). The bright spots are present on the continent related to the second object. The first continent is purple, corresponding to an activation of 0.25: when no information is available (i.e. the ESN makes a random guess). 
    \textbf{D.} The activation of the concept \enquote{Middle position for the object 1} (<middle object 1>). We can spot the moment where the keyword appears: when the color transition from bright purple to yellow on the first-object continent.}
    \label{fig:first_RSS_ESN}
\end{figure}

\begin{figure}
    \centering
    \includegraphics[width=\textwidth]{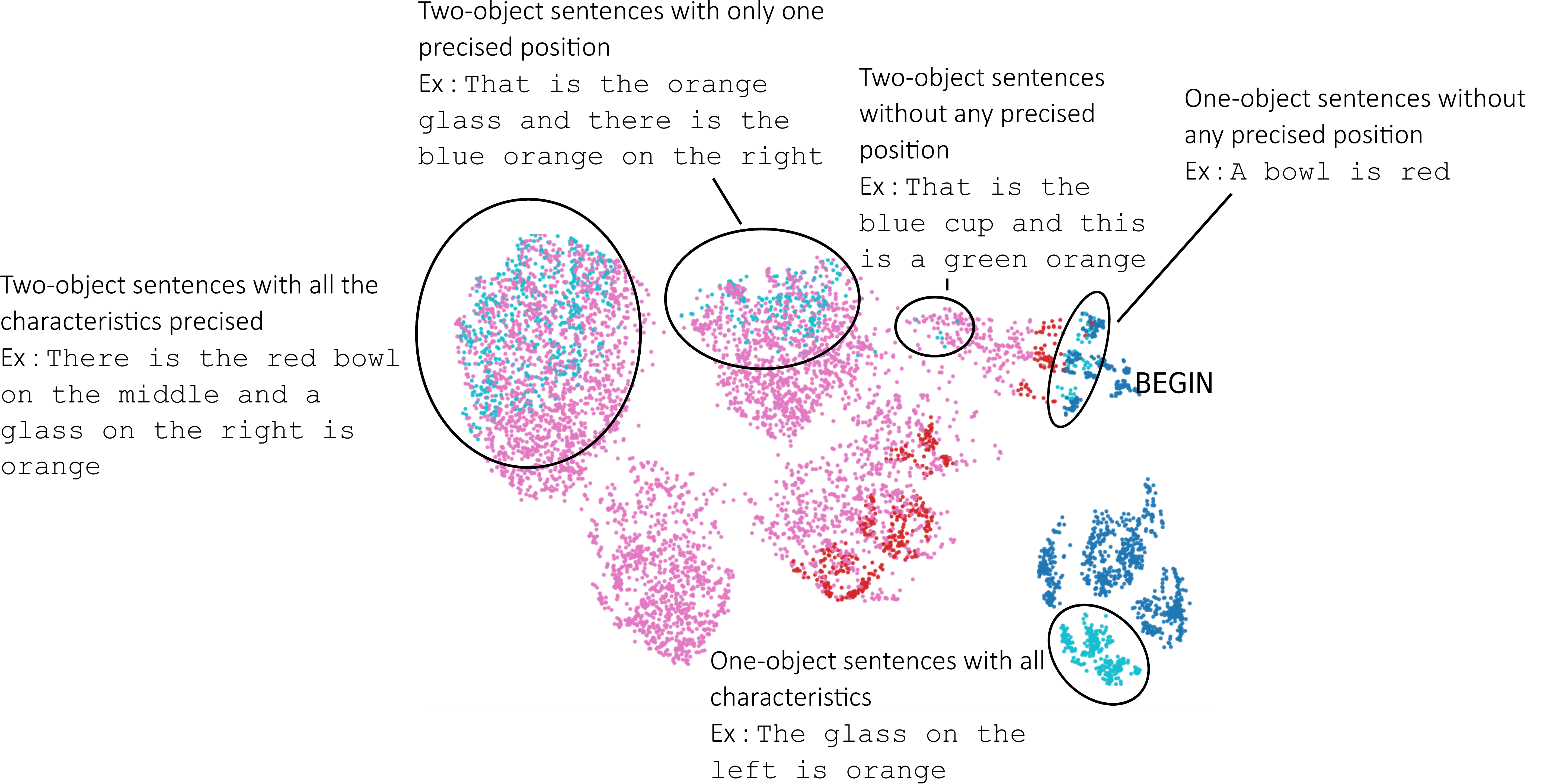}
    \caption{Description of the final states clusters in RSSviz for an ESN. Different clusters correspond to different sentences forms: with one or two object descriptions, with or without position precised.}
    \label{fig:RSS_ESN_structure}
\end{figure}

The results of RSSviz are shown in figure \ref{fig:first_RSS_ESN}. Since a sentence corresponds to a sequence of reservoir states, in RSSviz a sequence corresponds to a trajectory from the initial point after the word \enquote{BEGIN} to the final \enquote{END} word. Moreover, we can see that RSSviz faithfully renders the two-part structure of the sentences. Also, as shown in the figure \ref{fig:RSS_ESN_structure}, the final states are clustered in five main groups. Two groups in the continent related to the first object and three groups in the continent related to the second. One of the main reason for this clusterisation is the distance to the initial point.
The more words there are in the sentence, the more distant will be the reservoir state point from the initial \enquote{BEGIN} point. Similarly, the more words there are in a sentence, the more concepts (i.e. object, color, position) are described. Thus, the more distant we are from the initial point, the more concepts will be described. In other words, the distance to the initial point could be an approximated measure of its complexity (in terms of number of concepts).

Even if the position of the points are not related to the training of the ESN, we used a color scale to show the activation of a given output neuron. It shows us information both on how the ESN categorizes the reservoir states and how RSSviz represents the states corresponding to different categories. 
In figure \ref{fig:first_RSS_ESN}, some clusters are visible in the activation of the ESN. The ESN has learnt to categorize states that are close together to similar categories. Their closeness is due to similar words in the sentences, and thus, to similar reservoir states.

Moreover, we can analyse the activation to spot precise behaviors. For instance, in figures \ref{fig:first_RSS_ESN}.C and \ref{fig:first_RSS_ESN}.D, we can see that when no information related to the concept has been yet received, the activation correspond to a random guess (i.e. 0.25 because there are 4 possible answers, this value corresponds to bright purple dots in RSSviz). When a keyword related to the concept is seen, the activation jumps to 1 (bright orange dots). However if a word is related to the concept category (e.g. a word corresponding to the position is seen) but does not match the value searched by the output neuron (e.g. we visualize the activation of the neuron responding to the concept \enquote{left position} and the word \enquote{right} has been received), then the activation falls to zero (dark purple dots).
This behaviour arises because the ESN has been trained with Continuous Learning. It learnt to predict output values even when it doesn't have any information. This leads to an interpretation of the output activation of this ESN as a measure of its uncertainty. RSSviz can help us view where the ESN is certain or not about different concepts.

\subsection{Recurrent States Space Visualisation of a LSTM} \label{lstm-rssviz}

\begin{figure}
    \centering
    \includegraphics[width = \textwidth]{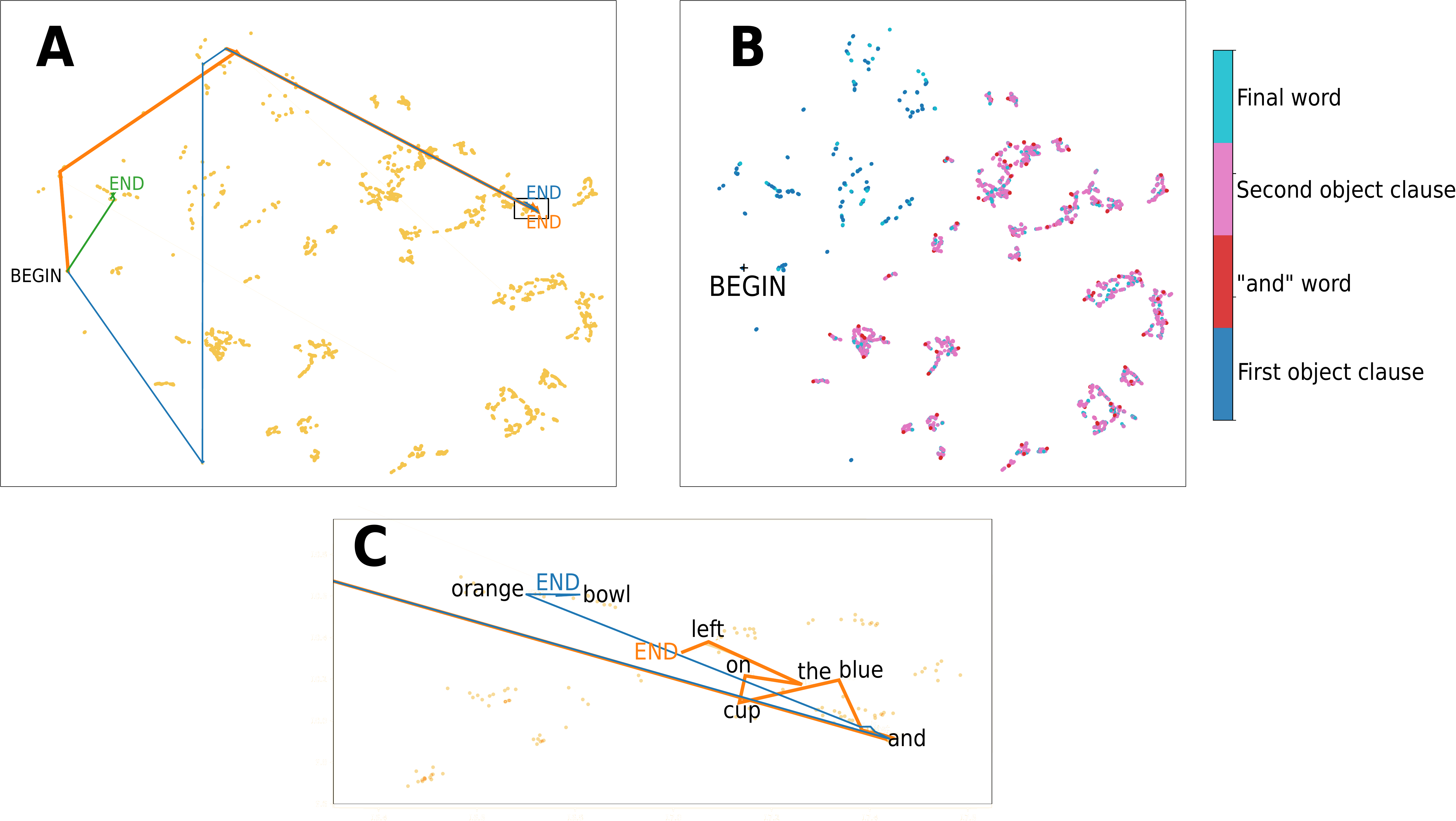}
    \caption{Structure of RSSviz for a LSTM. We can observe a hierarchical distinction between the processing of the first and second clause.
    \textbf{A.} Trajectories of 3 sentences in RSSviz. One sentence composed of the description of one object (green), two with two objects (blue and orange). These last two sentences describe the same first object but in different manner: \enquote{BEGIN the glass on the middle is green and that is a orange bowl END} (blue line), \enquote{BEGIN a green glass is on the middle and there is a blue cup on the left END} (orange line). 
    \textbf{B.} Visualisation of the states depending on their position on the sentence. We can observe an archipelago corresponding to the first object (dark blue) and one to the second object (in bright blue). The red dots, positioned on the outside of the islands in the second object archipelago, correspond to states where the word \enquote{and} is seen. We called them \emph{airports}. 
    \textbf{C.} Zoom on the end of the lines of the figure A. They end up on the same airport and after that, stayed in the same region. This is due to the fact that their first clause contain the same information even if they are structured differently.
    }
    \label{fig:RSS_LSTM_struct}
\end{figure}

The RSSviz representation of the LSTM is really different from the ESN one for two main reasons. First of all the recurrent states are learnt. Secondly, the state space %
dimension is 20 instead of 1000 for the ESN.
Because the inner representations of sentence parts are learnt, it means that they are optimized to tackle the task. That's why we can expect the recurrent states of the LSTM to be far apart according to the characteristics contained in the sentence. This makes the job easier for the linear classification operated by the final layer. In the RSSviz representation this is visible in figure \ref{fig:RSS_LSTM_struct}: the dots are sparsely clustered. Note that we didn't change the parameters of the UMAP embedding so we can make a comparison between the RSSviz of the ESN and the RSSviz of the LSTM. To introduce a geographical metaphor, if the RSSviz representation of the ESN is organized in \emph{continents}, the one of the LSTM is closer to to \emph{archipelagos}. Nonetheless, we can find common characteristics present in both RSSvizs. In the figure \ref{fig:RSS_LSTM_struct}.B, we can see a clear division between the states corresponding to the first and the second clause of the sentence. We distinguish the first object archipelago in dark blue and the second object archipelago in pink. The transition between these two archipelagos is made when the word \enquote{and} is seen (red dots). We called these special points \emph{airports}.

In figure \ref{fig:RSS_LSTM_struct}.A and \ref{fig:RSS_LSTM_struct}.C, we observe that the trajectory of a two-object sentence wonders in the first object archipelago from island to island during the first clause. Then, as soon as the word \enquote{and} is seen, it jumps to an airport on an island in the second object archipelago and ends its journey on this island, close to the airport.

In fact, this behaviour is not insignificant. It is due to the fact that the different islands and even each part of the different islands encodes for precise characteristics. In the zoom on the figure \ref{fig:RSS_LSTM_struct}.C, we can see that the trajectories of the two long sentences (blue and orange lines) end up on the exact same spot when the word \enquote{and} is received. This is explained by the fact that even if the first part of the sentence has a different structure, these sentences hold the same information about the first object. In RSSviz this information is encoded as a part of the space dedicated to the \enquote{first object: green glass middle} concepts combination, i.e. the position of the corresponding airport and the space around it. This shows that the LSTM learnt to focus on the features needed for the task, and not to take into account word order and non semantic words.

\begin{figure}
    \centering
    \includegraphics[width=\textwidth]{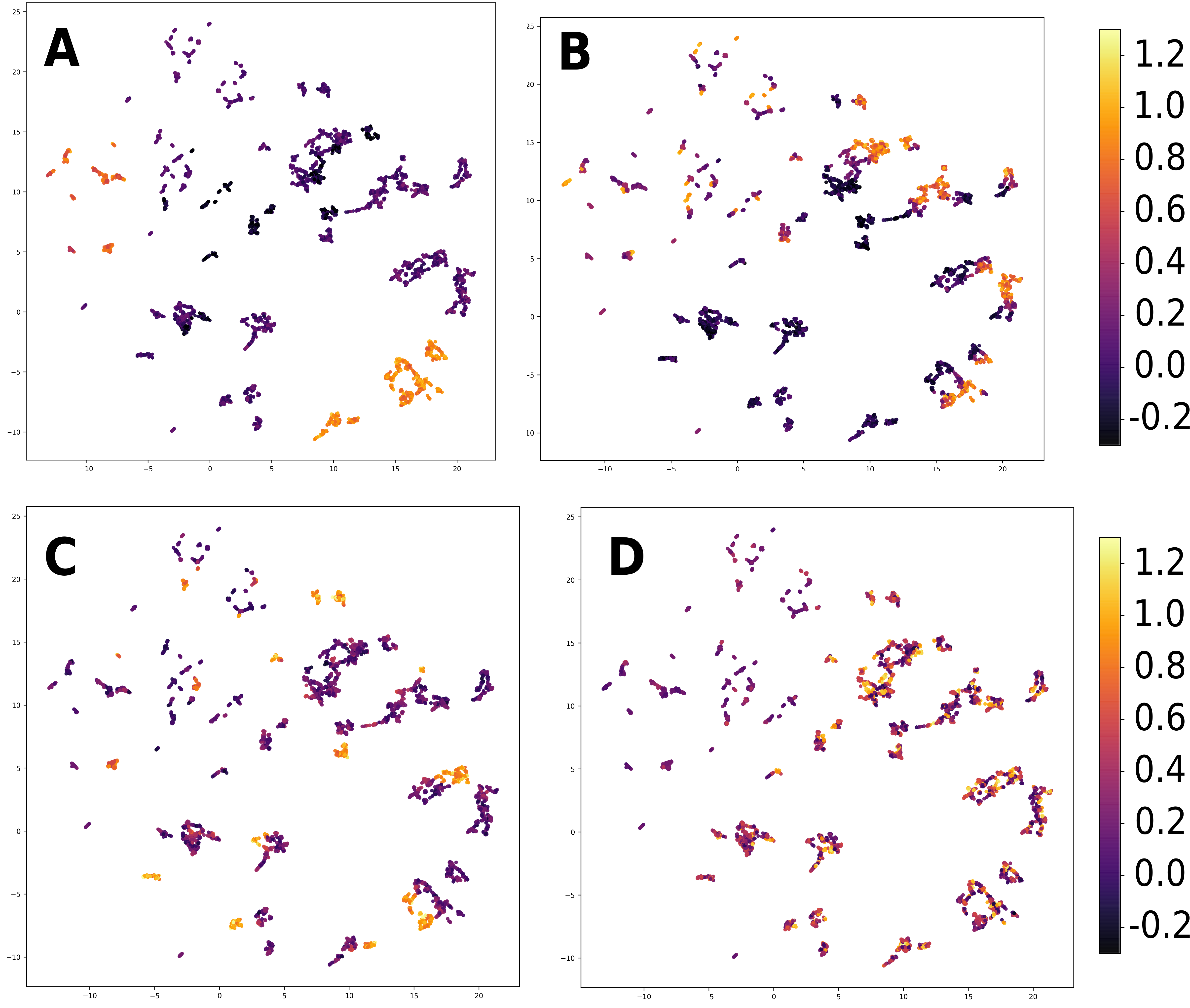}
    \caption{
    Activation of output neurons in RSSviz. %
    The less the concept is frequent, the smaller the clusters of its activation (in orange). From A to D the concept investigated becomes less and less frequent because the first object always has a category (e.g. "glass"), but not always a color, and less frequently a position. Finally, there is not always a description of a second object.
    \textbf{A.} Output activation for the first object category: \enquote{cup}.
    \textbf{B.} First object position: \enquote{left}.
    \textbf{C.} First object color: \enquote{blue}.
    \textbf{D.} Second object position: \enquote{middle}. All the characteristics related to the second object have this shape: high spatial frequency patterns in the archipelago of the second object.
    }
    \label{fig:RSS_LSTM_acti}
\end{figure}

This analysis can be pushed further by looking at at the distribution of the activations in RSSviz.
The results shown in figure \ref{fig:RSS_LSTM_acti} suggests that the places were each concept is activated are organized in widely different shapes. For the first object, the object category (fig. \ref{fig:RSS_LSTM_acti}.A) seems to divide RSSviz at the scale of whole islands, whereas the position and color of the object induce divisions at the scale of island parts (fig. \ref{fig:RSS_LSTM_acti}.B and \ref{fig:RSS_LSTM_acti}.C). The difference is striking with the concept related to the second object. They are all similar to figure \ref{fig:RSS_LSTM_acti}.D: they induce low scale, high spatial frequency division in the island of the second object archipelago.

\begin{figure}
    \centering
    \includegraphics[width=\textwidth]{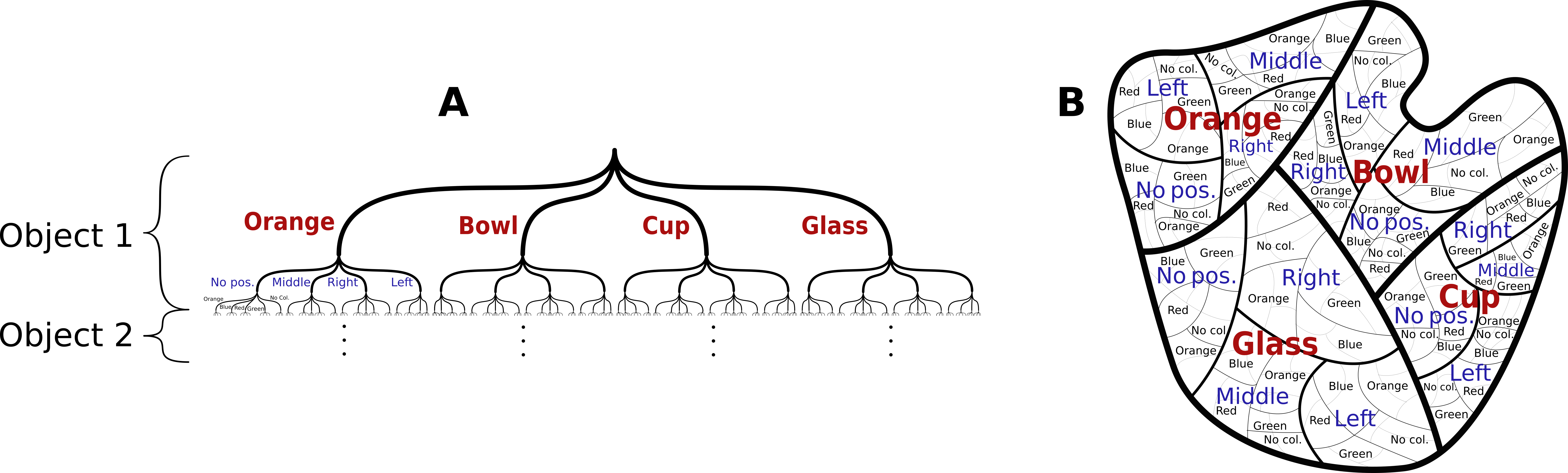}
    \caption{
    Schematic representation of the fractal encoding learnt by the LSTM.
    \textbf{A.} The LSTM has to find the right branch in the categorisation tree to recognize the characteristics for both objects. Because an object can have no position, or no color, there exists corresponding branches. Even if these branches don't directly correspond to the activation of a single neurons, the LSTM has to acknowledge them in order to change its output accordingly.
    \textbf{B.} The shape of the activation in RSSviz suggests that the LSTM learnt a fractal representation of the abstract categorisation tree depicted in A. The spatialisation of the tree can be described with two rules: (i) Each node correspond to a region in the space (ii) Each node has its region located \emph{inside} the region of its father without overlapping with other son's region.
    }
    \label{fig:schematic_rss_lstm}
\end{figure}

We interpret these different scales of classification as a form of spatialisation of a categorisation tree. As illustrated in figure \ref{fig:schematic_rss_lstm}.A, the model has to chose the right branch of the tree to succeed in doing the task. 
Its inner representation was optimized to encode spatially this tree. We can describe the process of tree spatialisation with two rules: (i) each node correspond to a region of space. (ii) All the sons of a node have their corresponding region inside the region of their father without overlapping.
Those simple rules lead to a fractal encoding of object characteristics, as outlined in figure \ref{fig:schematic_rss_lstm}.B.

Nonetheless, this interpretation has some limits. We can understand why the first object characteristic has an influence: every object have a type (i.e. category) but not necessarily a position nor a color.
However, the hierarchy between position and color in the classification tree is unclear: it is possible that objects without color are more frequent than object without position. In all cases, we can think of it as fuzzier than what is depicted in \ref{fig:schematic_rss_lstm}.A. Notwithstanding, the hierarchy between the first and the second object characteristics is significant. It makes sense because when the description of the second object begins, all the information related the the first object is known.
That explains the high frequency pattern in the activation of concepts related to the second object. In fact, the region of the space corresponding to the node at a depth 3 in the tree in \ref{fig:schematic_rss_lstm}.A are the airports and the region around them, visible in figure \ref{fig:RSS_LSTM_struct}.C. 

These findings are a generalization of what we have seen in figure \ref{fig:cells_LSTM} were we measured that semantic words caused more change in cells than non semantic words. Here, we show that this is in fact caused by an underlying rich structure encoding for the sentence features.
This structure also explains why we can observe a jump in output activation as soon as a corresponding keyword is seen (fig. \ref{fig:acti_plots}.(c)). When a keyword is seen, the trajectory of the sentence jumps into the corresponding region of the space. The final layer has learnt to associate this region of the cell space with the corresponding activation. Moreover, since the trajectory of the sentence will stay in this region, the concept will be activated for the rest of the sentence.

This fractal structure for encoding sequence was investigated in \cite{carmantini2017modular} to make a bridge between symbolic and a vectorial space in the context of neural network and non linear automaton. This encoding emerges here naturally as a result of the learning of the LSTM, similarly to the \textit{Fractal Neural Networks} developed by Tabor \cite{tabor2003learning}. %
This structure also provides another way to look at the memory mechanism in LSTM and in RNN in general. In this visualisation, the memory is visible spatially. Remembering a piece of information means staying in a part of the space corresponding to a feature seen in the past but no more visible/accessible. This spatialisation of the memory is similar to the one in dynamical automata, an automaton with a state formed by real numbers. We can design update rules for their state that enable them to recognized any context free grammar \cite{tabor2000fractal}. These update rules then organize the computation of a given string by the dynamical automaton in a continuous space, as the LSTM. Because of the recursive nature of context free grammars, the trajectories of the dynamical automata often take the form of fractal. The visualisation of the organisation of the hidden cells of the LSTM suggests that the LSTM organizes its recursive computations in a similar manner as the hand-made dynamical automata. It seems to be a natural way to organize the computation of the huge number of different sequences in a bounded space.
Indeed, we think that this fractal structure emerges here mainly because of the recursive nature of the task: the identification of the information related to the second object is pretty much the same task as for the first object.
Nonetheless, this structure is likely not specific to this task. It could emerge as a way to handle any symbolic sequence classification task that presents concatenation of similar patterns. More work on other tasks is needed to experiment these type of analysis in other context.

\section{Practical use cases for the Recurrent States Space Visualisation}
\label{sec-pract-use-cases-RSS}

In this section we will investigate how RSSviz can be useful in a practical manner when designing RNN. We will primarily focus on ESN for this part. More work is needed to see if the following points can also be applied to LSTM or other models.

\subsection{RSSviz can help us spot errors}

As discussed in section \ref{sec-RSS_ESN}, the activation of output neurons visualised with color scale in RSSviz are, in a sense, a picture of the uncertainty of the model. We can exploit this observation to spot places were the ESN seems uncertain about its prediction. 

\begin{figure}
    \centering
    \includegraphics[scale = 0.42]{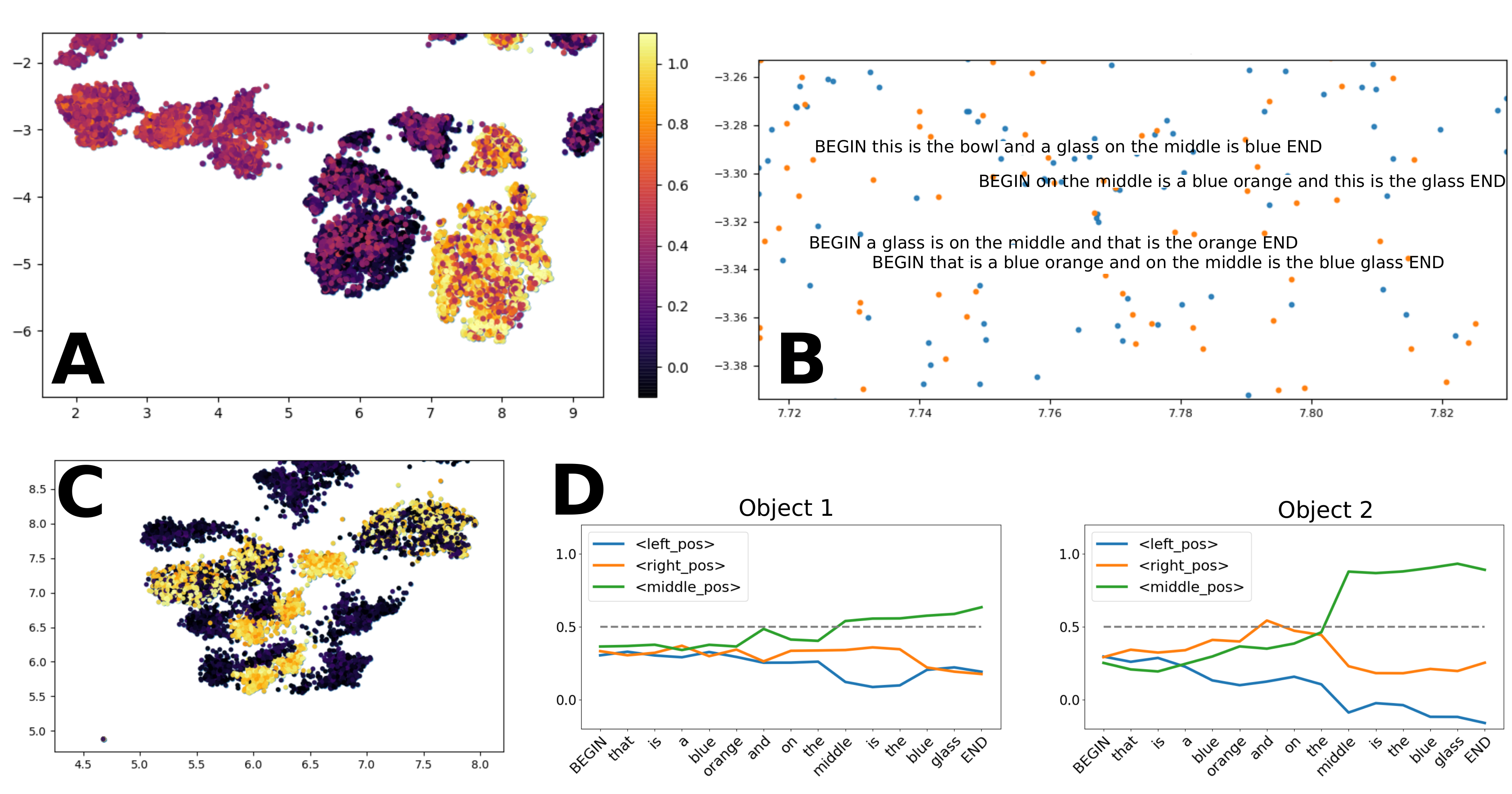}
    \caption{RSSviz can help us identify difficult sentences.
    \textbf{A.} Activation of the concept \enquote{Middle position for object 1}. RSSviz made with reservoir states gathered on 10,000 sentences. The part shown here is a zoom on an island with different color close together. It reflects the uncertainty of the ESN on these states. 
    \textbf{B.} The sentences corresponding to the reservoir states on the island of the figure A. The color of the points is not related to activation: orange points are final states and the blue ones, normal states. 
    \textbf{C.} A zoom from the same RSSviz as the figure A. A typical island were the concept \enquote{Middle position for object 1} is activated. Contrary to the figure A, the color of the cluster is homogeneous. It reflects the confidence of the ESN in its prediction. 
    \textbf{D.} Output activation graph for the sentence \enquote{BEGIN that is a blue orange and on the middle is the blue glass END}, identified in figure B. The ESN believed that the concept \enquote{middle} was attributed to both objects. Its representation was valid but not exact. Exploring RSSviz enables us to discover this sentence, difficult for the ESN.}
    \label{fig:explain_error_ESN_RSS}
\end{figure}

We created a RSSviz of an ESN with more details: we gathered the recurrent states on 10,000 sentences instead of 1,000.
As shown in figure \ref{fig:explain_error_ESN_RSS}, in this RSSviz we plotted the activation of the \enquote{middle position for object one} output neuron.
Then, we spotted a cluster that seemed less bright than the other and that contained different colors packed together. This suggests that the model struggles to categorize these reservoir states because of their closeness. When we search for the corresponding sentences, we found sentences with the word \enquote{middle} related to the first or to the second object. Because the word \enquote{middle} was close to the word \enquote{and}, the ESN had a hard time determining to which object the concept \enquote{middle} is related to. Moreover, to make the sentence more difficult, the other object had no position precised. It leads the ESN to predict, as shown in figure \ref{fig:explain_error_ESN_RSS}.D, that both objects were on the middle. This is a valid but not exact representation.

Thanks to RSSviz, we have been able to spot what are the difficult examples for this particular ESN. This method to spot errors is akin to the way adversarial examples were spotted thanks to Activation Atlas in \cite{carter2019activation}. In the future, this could be useful to spot the flaws of certain model and to invent new models that correct these limitations.

\subsection{We can create dynamic RSSviz to understand how the learning takes place}

\begin{figure}
    \centering
    \includegraphics[width = 0.9\textwidth]{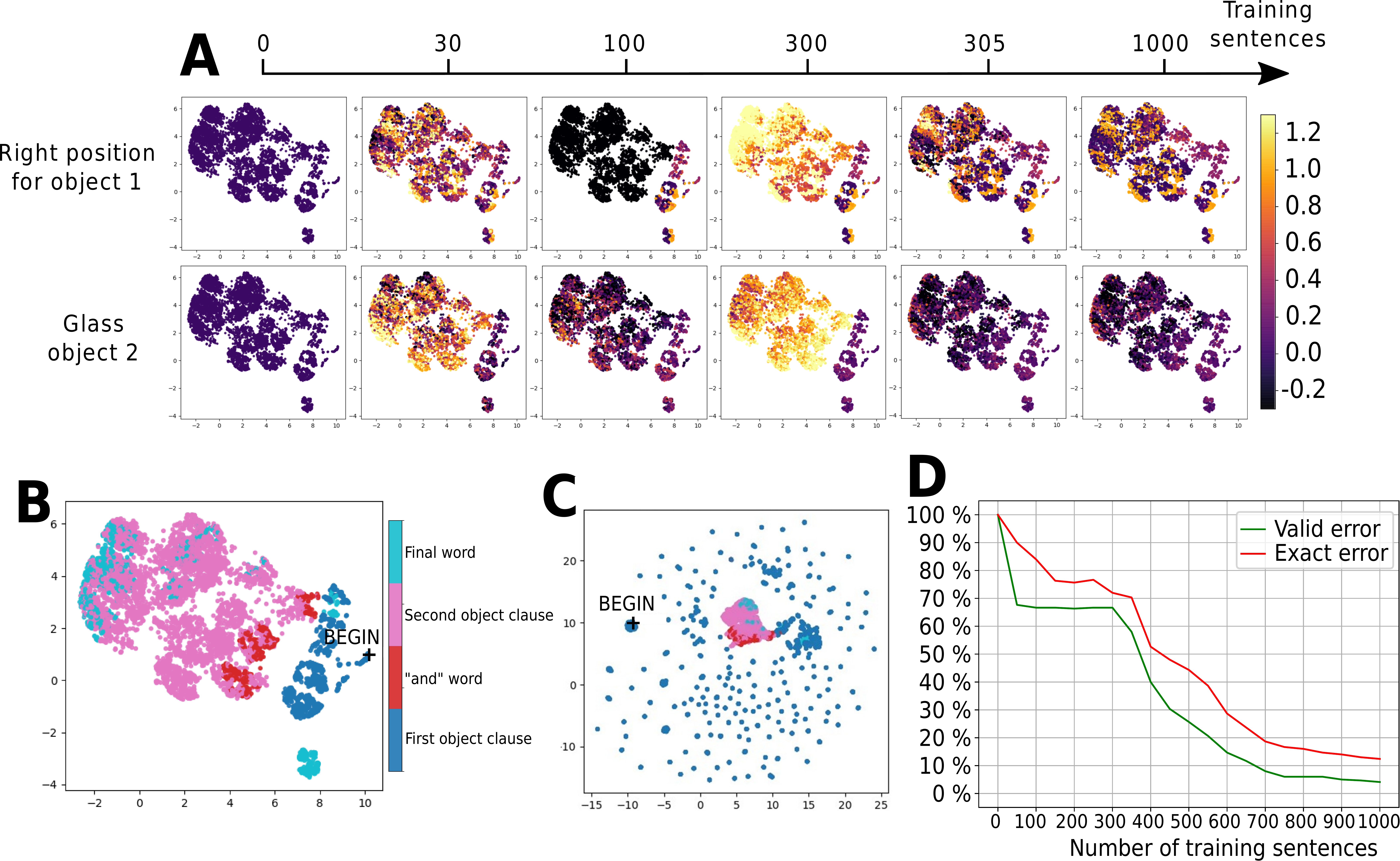}
    \caption{We can view the learning process of the ESN thanks to RSSviz. %
    During the first 300 training sentences, the ESN learnt to categorize the sentence describing one object and only after the training sentence \#300, it began to successfully handle the two-object sentences.
    \textbf{A.} Evolution of the activation of two output neurons during learning. The images are not regularly spaced in time: only key changes are depicted. The color scale is fixed for all the pictures, saturation is applied for values outside the range of the color scale. After the sentence 300, a rapid change happened and the ESN learnt to categorize two-object sentences. 
    \textbf{B.} Structure of RSSviz of the ESN. 
    \textbf{C.} RSSviz created without removing duplicate reservoir states. We can see that the states corresponding to the first object are much more present and thus take more space in this version of RSSviz. 
    \textbf{D.} Evolution of the error on a validation set composed by 300 one-object sentences and 700 two-object sentences. The learning happens by stages: the one-object sentences first (when the error plateaus to 0.7) and after the sentence 300, the two-object sentences.
    }
    \label{fig:learning_process_rss_esn}
\end{figure}

In the ESN used here, the reservoir isn't influenced by the learning process. It means that we can create the points of RSSviz before training, their positions will not change during the course of the learning. However, the readout weights will be optimized to learn the task after each training sentence. This means that activation of outputs neurons will change during learning. We can visualize how they learn to select the pertinent states by creating colored RSSviz at regular time interval during the learning.
We created a video showing how the activation on RSSviz changes during learning for several output concepts. The video is available \href{https://www.youtube.com/watch?v=mO90C8oP7u0}{ \color{blue}here}\footnote{https://www.youtube.com/watch?v=mO90C8oP7u0}. A temporal representation is shown in figure \ref{fig:learning_process_rss_esn}.A where key images of the video are shown. We can see that the learning process is not uniform in time. Indeed, the activation on the first object continent is quickly stabilized but the second object continent takes much more time to be learnt. Moreover, around sentence 300, we can see a brutal change in the activation on the second object continent happening in just five training sentences. This break point is observed simultaneously for all the concepts. After sentence 305, no major changes are seen on the second object continent until the end of the training.
In figure \ref{fig:learning_process_rss_esn}.D we observe that the breakpoint moment corresponds to when the ESN succeed in categorizing two object sentences: the valid error begin to fall below 0.7.

 It makes sense that one-object sentences are learnt quicker. Even if they are not in majority in the training set, because we use continuous learning, the first parts of two-object sentences act as a training for one-object sentences. In figure \ref{fig:learning_process_rss_esn}.C, were RSSviz was made without removing duplicated reservoir states, we can see how all the states from the second object clause are tightly packed together despite holding widely different information. There is just more states corresponding to the first object. In some sense, this figure shows more faithfully the dataset the ESN has to learn: it also account for the statistical distortion included in the dataset. However it is hard to distinguish structure in the second object continent of figure \ref{fig:learning_process_rss_esn}.C, that's why we kept only unique reservoir states in the other figures.

These dynamical visualisation can also be interesting to create developmental models and to track how learning is happening. It can be useful to evaluate the biological plausibility by comparing to how learning happens in a brain for example.
However, more work needs to be done to apply these visualisations to LSTM were both the activation and the position of the point move at the same time  during the learning process.
Some attempts in this direction have already been made by Rakesh Chada to visualize the evolution of LSTM word embedding on hate speech recognition task. The results can be found in this blog post \cite{understanding-lstm-embedding}.

\subsection{RSSviz can be a way to understand the influence of hyper-parameters}

The choice of hyper-parameters for the reservoir is currently an active field of research. As we saw in the previous parts with the LSTM and the ESN, RSSviz can be used to understand the memory abilities of a model. It can also depicts how the model represents in its inner activation the features of the input sequence. In fact, these abilities are the ones investigated when designing a reservoir for an ESN. 
That's why, in this section we will investigate how the structure of RSSviz changes when we deviate one hyper-parameter from the optimal combination. 
Because the goal of this section is to optimize performances, we chose here to work with the ESN trained with final learning, were it performed the best.

\subsubsection{Influence of the leak rate}

\begin{figure}
    \centering
    \includegraphics[width = \textwidth]{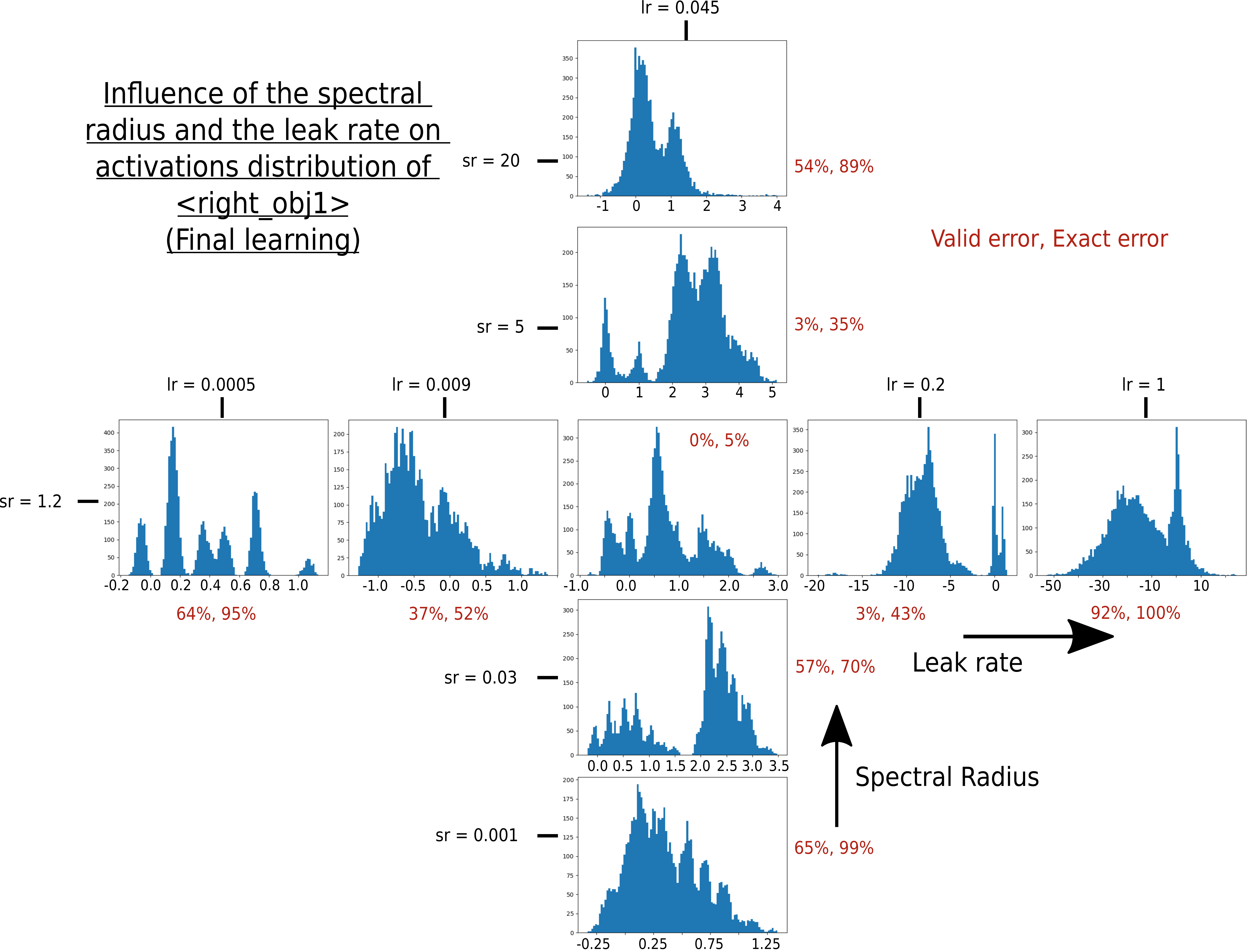}
    \caption{Distribution of output activations for the neuron corresponding to the concept \enquote{Object 1 on the right}. The histograms are shown for several reservoir hyper-parameters combination where one hyper-parameter deviated from the optimal. Notice the changing scale in the magnitude of the activations: high leak rate leads to chaotic divergent behaviours. Low leak rate and low spectral radius cause multi-modal  distribution between 0 and 1 with sharp peaks due to a processing of the information more linear. }
    \label{fig:cross_plot_hist_FL}
\end{figure}

\begin{figure}
    \centering
    \includegraphics[width = \textwidth]{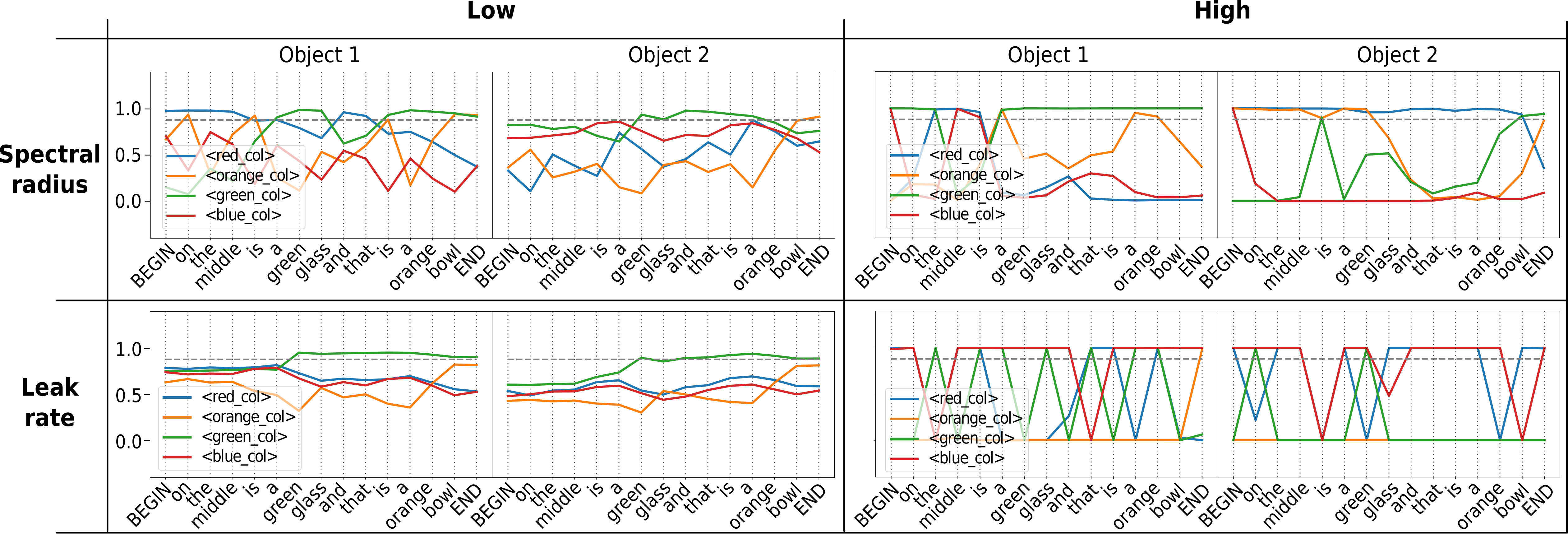}
    \caption{Output activation plots for the color of both object. High leak rate and high spectral radius leads to chaotic behaviours whereas when they are low, the ESN presents linear behaviour and is not able to differentiate the two objects.
    The output are shown after the application of the sigmoid function $\sigma : x \mapsto \frac1{1 + {\rm e}^{-3x}}$ to get bounded values. Four hyper-parameter combinations are shown where one hyper-parameter deviated from the optimal. Hyper-parameters values: high leak rate: 1.0, low leak rate: 0.0005, high spectral radius: 20, low spectral radius: 0.001}
    \label{fig:acti_plots_hp}
\end{figure}

\begin{figure}
    \centering
        \begin{subfigure}{\textwidth}
        \centering
         \includegraphics[width=0.9\textwidth]{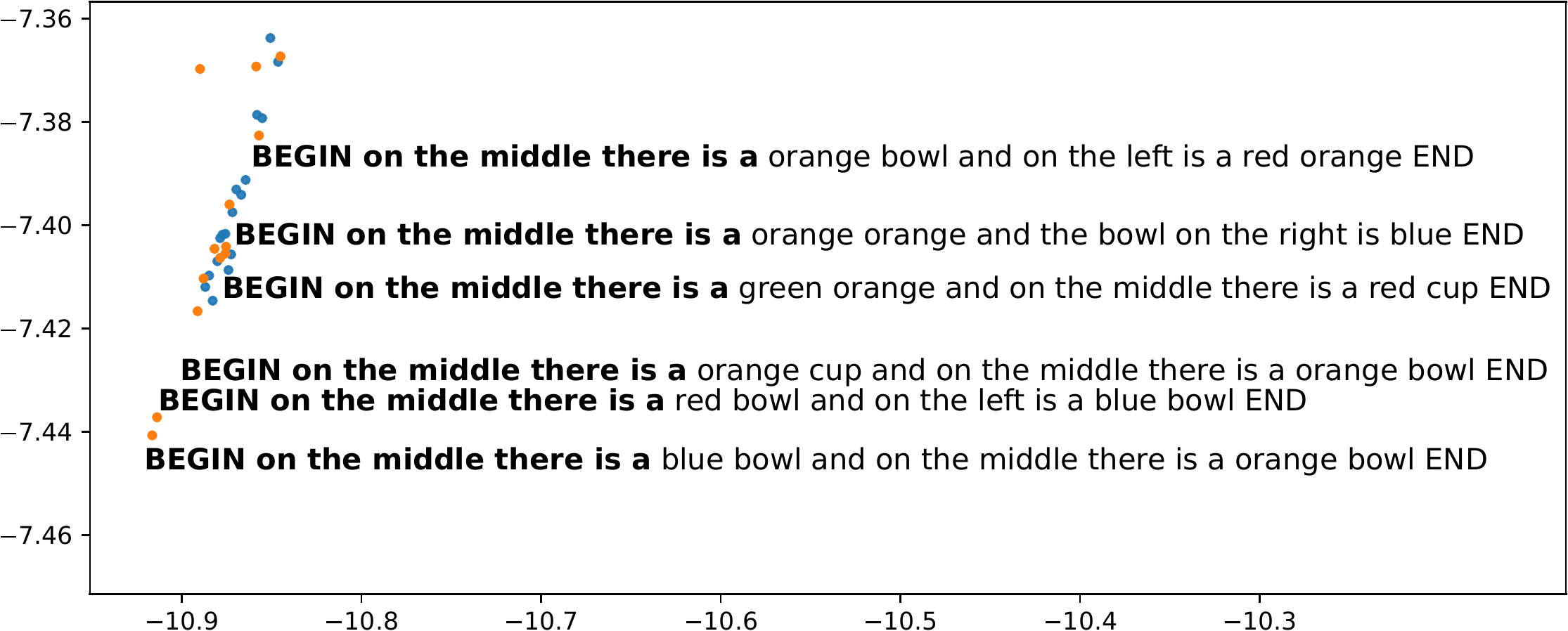}
         \caption{For high spectral radius (sr=20), the states are clustered according to their first words: here, they all begin by the words \enquote{BEGIN on the middle there is a}.}
    \end{subfigure}
        \begin{subfigure}{\textwidth}
        \centering
         \includegraphics[width=0.9\textwidth]{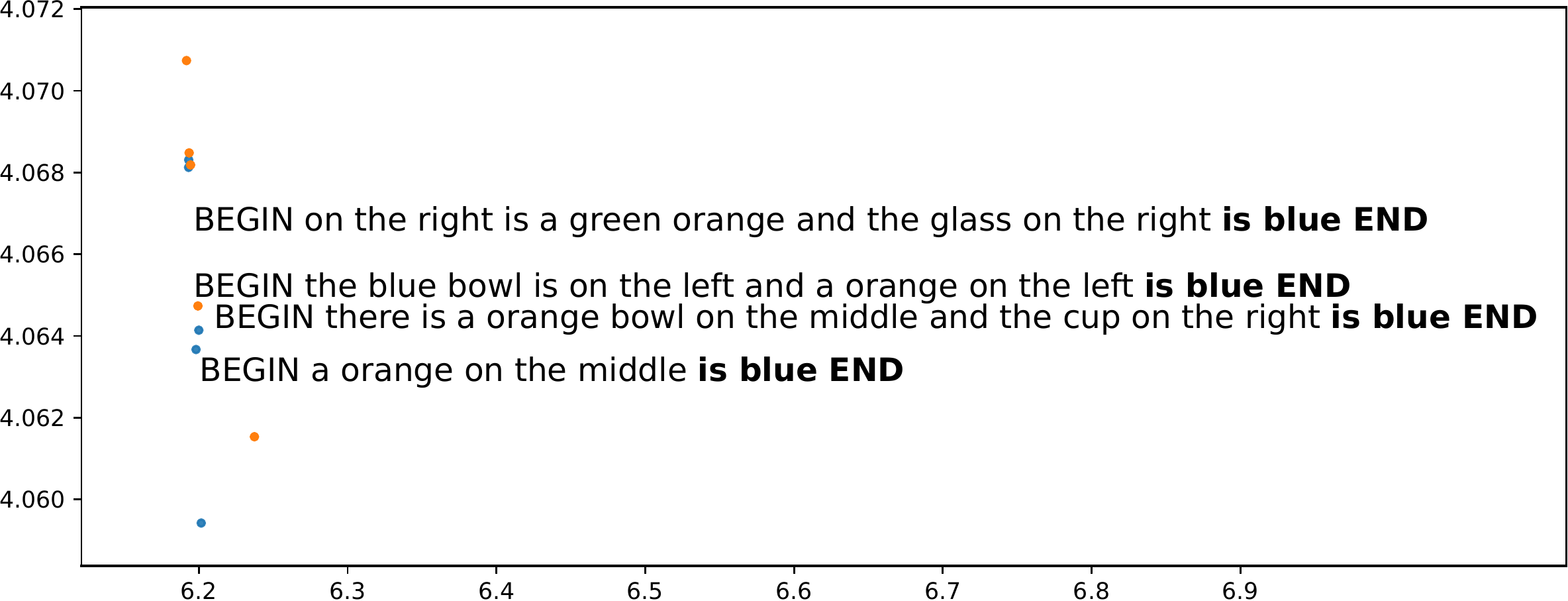}
         \caption{For high leak rate (lr=1), it's the opposite. The final states are organized according to their last words, here, \enquote{is blue END}.}
    \end{subfigure}
    
    \caption{ For high spectral radii and high leak rates the final states are organized in clusters determined by the first or last words of the sentence.}
    \label{fig:final_states_high_sr_high_lr}
\end{figure}

As show in equation (\ref{eq_reservoir}), the leak rate controls how much we keep from the previous state to compute the new one. It directly controls a trade-off between memory and new information.
For high leak rate (close to 1), the reservoir state is highly determined by the new piece of information received. In RSSviz, this leads to a sparse space with clusters organized by similar final words, as shown in figure \ref{fig:final_states_high_sr_high_lr}.B . This structure makes it difficult for the ESN to exploit information about the first object. The memory of the reservoir is highly limited. In fact the behaviour of the reservoir is here chaotic. As visible in figure \ref{fig:acti_plots_hp}, it seems hard for the ESN to extract any meaningful information from it. This is why we have an error this high.
For low leak rate (around 0.005), it is the opposite. The new reservoir state is really close to the previous one. It means that the reservoir has a important memory. However, its information processing abilities are limited: as shown in figure \ref{fig:acti_plots_hp}, the ESN is unable to differentiate between characteristics related to the first and to the second object. This is why it gets around 64\% in valid error: it was able to output a valid representation for all the one-object sentences (30\% of the test set) but was nearly always not valid on two-object sentences.
In fact it seems to behave like a simple automaton: when a keyword is seen, the corresponding concept activates and the others are inhibited for both objects. This leads to output activation graphs in figure \ref{fig:acti_plots_hp} that are organized in steps. This automaton-like behaviour also causes a distribution of activation shown in figure \ref{fig:cross_plot_hist_FL} composed of several sharp peaks. They correspond to the different combination of inhibition and activation. For example, the first peak at -0.25 correspond to a concept inhibited twice, the peak around 1.1, to a concept activated twice, the peak around 0.6 to the concept activated and then inhibited, and so on. 

Remarkably, even if we used Final Learning (FL), the intermediate outputs are interpretable for this hyper-parameters combination. This behaviour seems to illustrate the compromise between memory and non linearity in reservoir, notably investigated in \cite{verstraeten2010memory}. Because this reservoir has a great memory, its abilities to process information in a non linear manner are limited. That's why this linear, interpretable automaton-like behaviour emerges.

\begin{figure}
    \centering
    \includegraphics[width=\textwidth]{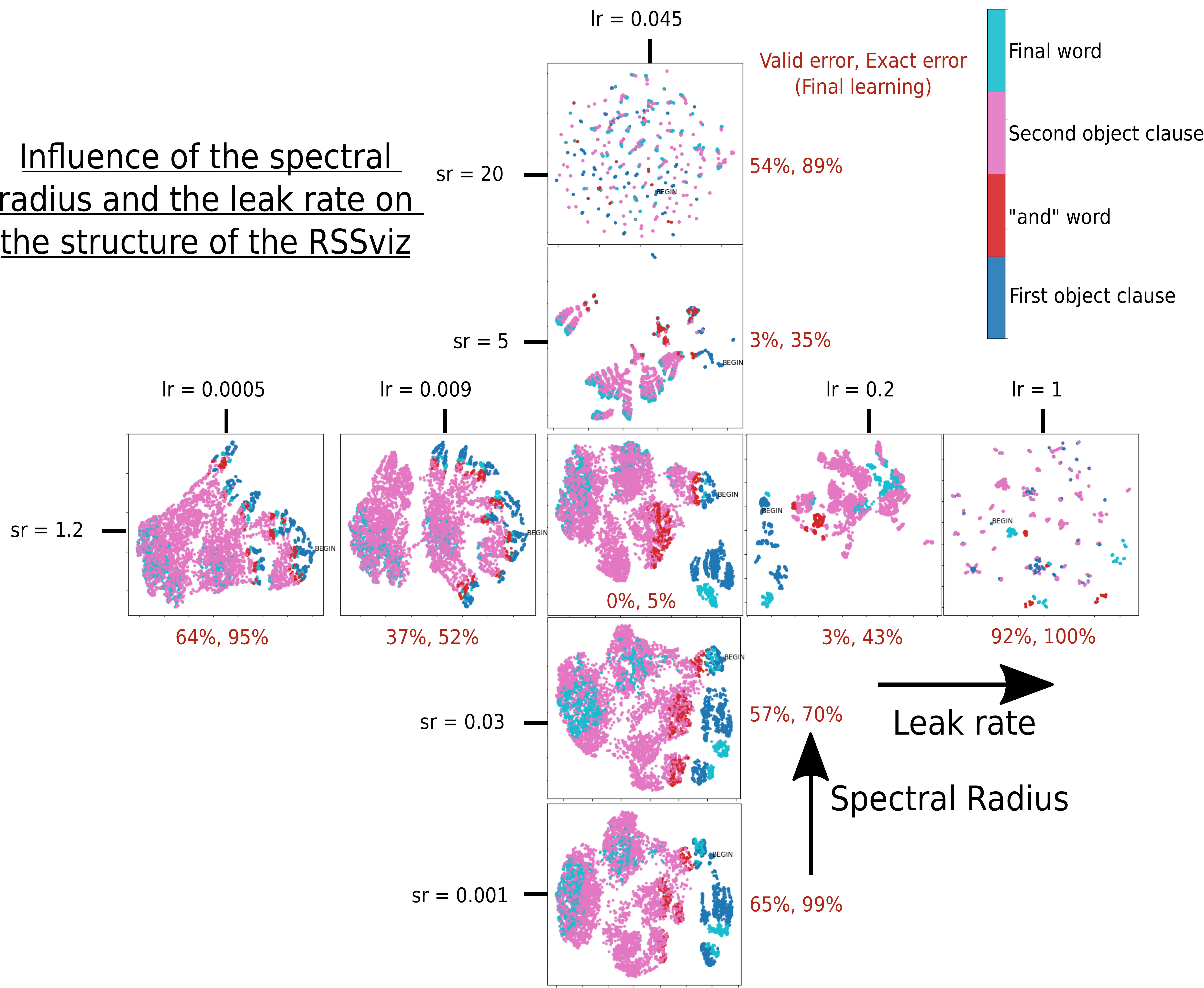}
    \caption{The structure of RSSviz for different reservoir hyper-parameters when one hyper-parameter is deviated from the optimal combination (in the center).
    The closer the hyperparameters are from chaos (high leak rate and high spectral radius), the more the RSSviz are fragmented.
    The text in red is the error of the ESN trained on the corresponding reservoir with FL. }
    \label{fig:cross_plot_struct}
\end{figure}

\subsubsection{Influence of the spectral radius}

In \cite{lukovsevivcius2012practical}, a high spectral radius is said to be useful for task requiring long term memory.
We found results coherent with this affirmation. For high spectral radius (around 20), as shown in figure \ref{fig:final_states_high_sr_high_lr}, we found that final states were clustered according to the first words of the sentence. This lead, in figure \ref{fig:cross_plot_struct} to a sparse space with islands determined in majority by the first words. In a sense, it's the opposite effect of high leak rate. 
With low spectral radius, we found output activations distribution with several peaks but they were less sharp than the ones with low leak rate (fig. \ref{fig:cross_plot_hist_FL}).
In figure \ref{fig:acti_plots_hp}, we can see that the ESN is still unable to differentiate between the first and the second object characteristics. This is  maybe due to a lack of non linear computation.

\section{Discussion}
\label{sec-discussion}

In this work we compared LSTM and ESN on Cross-Situational Learning. Those are two popular and successful RNN architectures used to handle sequential data. We show that, on this task, the ESN was far less expensive in computation than the LSTM to achieve slightly worse results on the 4-object dataset. It outperformed the LSTM for datasets more challenging. As what has been already found on other tasks, the ESN appears as an efficient model for language processing. However, the hyper-parameters of the ESN were found by extensive random search in \cite{juven2020cross}. %
Whereas for the LSTM, we did almost no hyper-parameter search to find a model performing well on the 4-object dataset. In fact the error was so low that we couldn't hope to do much better after a hyper-parameter search. To account for these disparity in the model designing process, further work could create a comparison in \emph{practical time}. In other words, we could account for all the computation time spent from the discovery of the task to the creation of a model performing below a fixed error level. This way, we include the computation time used for the hyper-parameters search in the comparison.

We also conducted analysis of inner state activation for both models but with different method. For the LSTM we analyzed the evolution in activity of all its cell units during the processing of the sentence. Whereas for the ESN, we used a form of feature selection: we analyzed the activity of reservoir units the most connected to a given output neuron. Even if these two models are deeply different, we were able to point out common inner working. In both cases we were able to identify recurrent units keeping track of the sentence structure. This behaviour was optimized in the case of LSTM, while the reservoir first creates random features before selecting among them the \enquote{useful} ones for the task.

To get a more holistic view at the inner working of both models, we created 2D visualisations of the high dimensional structure of the recurrent state space (RSSviz). To this end, we used UMAP to efficiently reduce the dimension while preserving the space structure. This method gives us hindsight at how the LSTM optimized its inner state to better tackle the categorisation task: we observe the emergence of a fractal spatialisation of the cell states. This phenomena is close to what observed Tabor in \cite{tabor2003learning}. By training a  \textit{Fractal Learning Neural Network} to recognized a context free grammar, he observed a fractal topology in the hidden units that organized recursive computations. These results are similar to what we showed here. Nonetheless, contrary to Tabor who used a specially designed network trained with hill climbing, we used a standard LSTM trained with a BPTT-based technique. We observed a fractal structure after the application of a non-linear dimension reduction, whereas Tabor directly analysed hidden units.
Contrary to the work of Tabor, this structure emerges in a practical context. The main point of this contribution was not about fractal structure in hidden units, in fact we didn't expected these observations. This suggest that fractal topology could be a useful tool to understand sequential data processing in RNNs. However, as did Tabor, our work relies on a context free grammar. Future work could investigate whether fractal organisation of the internal state of RNN can emerge after training on real world data.

For the ESN, RSSviz was a way to visualize how the readout matrix divide the reservoir states space (with hyperplanes) to generate the right output activation for each unit. Moreover, these tools helps us uncover what were the difficult examples for the ESN. It also enables us to create animated visualisation of the ESN learning process. These videos showed that the learning happened in stages: the sentences describing one object were learnt before sentences with two objects. More work is needed to investigate if similar phenomenon can be observed in LSTM.

We also showed that RSSviz can be used to investigate the effect of reservoir hyper-parameters. Even if we examined only two reservoir hyper-parameters: the spectral radius and the leak rate, we were able to identify trade-off between non linearity and memory in the reservoir. However, we still lack a whole understanding of every hyper-parameters. Moreover, this study only investigate the deviation of one hyper-parameter from the optimal combination. We could investigate how several hyper-parameters interact together. We used primarily qualitative analysis for this part, in the future, we could investigate how to introduce quantitative analysis with clusterization metrics for instance. Nonetheless, this new way of visualising the reservoir hyper-parameters is very promising because it can be created before any learning procedure. This could be a tool to visualize if the reservoir possesses some desired properties. If we are able to quantify the quality of a reservoir thanks to its RSSviz, we could be able to accelerate hyper-parameter search. Indeed, this is currently a computationally demanding process in designing high-performance ESN. For high dimensional ESN, the UMAP transformation is faster than training the ESN.

Finally, RSSviz could also help to enhance the \enquote{exploitation} of ESN internal representations in a unsupervised or supervised way. In an unsupervised perspective, it could help to better understand the effect of of homeostatic rules, such as \emph{Intrinsic Plasticity} \cite{steil2007online}, to adapt the recurrent weights of the ESN. In an offline supervised perspective, it could help design new ways of training the output layer given observations of the spread of internal states, and which mechanism could enhance some distribution of these internal states, from the output perspective. In a online supervised perspective (such as FORCE learning), studying the evolution of RSSviz color maps can help to design new online learning algorithms.
Moreover, it could also help to understand the influence of the feedback when outputs are fed-back to the reservoir. Given that feeding back outputs into the reservoir can reduce the dimensionality of the itnernal states (especially for symbolic tasks) \cite{pascanu2011neurodynamical, hoerzer2014emergence, strock2020robust}, studying this effect more deeply with RSSviz would provide more insights on the complex influence of the feedback in general. We believe that studying the combination of both unsupervised rules and feedback could lead to new learning rules for ESNs.

The breath of these conclusions about RSSViz is quite limited: we investigated only two architectures on a single task. Further work is needed to explore how these visualisations apply to other RNN such as GRU and on other tasks with fuzzier structure in datasets. For future perspective, this method could be useful to spot were are the particular flaws of a given model. We could use this information to make relevant proposal of new model architectures.

In a more general way, we think that inventive visualisation could be used in the mainstream workflow when designing an RNN or even any artificial neural network. Instead of keeping track of a little number of narrow metrics, visualization of ANN could leverages our visual abilities in pattern recognition. This paradigm could enable more informed decision and hindsight on the role of hyper-parameters and model structure.

\section{Acknowledgments}

We greatly thank Alexis Juven and Thanh Trung Dinh for the development of previous source code on reservoir implementation with FORCE learning and the cross-situational task. We are also grateful for fruitful discussions with them.

\FloatBarrier
\bibliographystyle{unsrt}  
\bibliography{main}  %

\begin{thebibliography}{10}

\bibitem{hochreiter1997long}
Sepp Hochreiter and J{\"u}rgen Schmidhuber.
\newblock Long short-term memory.
\newblock {\em Neural computation}, 9(8):1735--1780, 1997.

\bibitem{rumelhart1986learning}
David~E Rumelhart, Geoffrey~E Hinton, and Ronald~J Williams.
\newblock Learning representations by back-propagating errors.
\newblock {\em nature}, 323(6088):533--536, 1986.

\bibitem{ma2015long}
Xiaolei Ma, Zhimin Tao, Yinhai Wang, Haiyang Yu, and Yunpeng Wang.
\newblock Long short-term memory neural network for traffic speed prediction
  using remote microwave sensor data.
\newblock {\em Transportation Research Part C: Emerging Technologies},
  54:187--197, 2015.

\bibitem{sak2014long}
Hasim Sak, Andrew~W Senior, and Fran{\c{c}}oise Beaufays.
\newblock Long short-term memory recurrent neural network architectures for
  large scale acoustic modeling.
\newblock 2014.

\bibitem{sundermeyer2012lstm}
Martin Sundermeyer, Ralf Schl{\"u}ter, and Hermann Ney.
\newblock Lstm neural networks for language modeling.
\newblock In {\em Thirteenth annual conference of the international speech
  communication association}, 2012.

\bibitem{steil2007online}
Jochen~J Steil.
\newblock Online reservoir adaptation by intrinsic plasticity for
  backpropagation--decorrelation and echo state learning.
\newblock {\em Neural Networks}, 20(3):353--364, 2007.

\bibitem{jaeger2004harnessing}
Herbert Jaeger and Harald Haas.
\newblock Harnessing nonlinearity: Predicting chaotic systems and saving energy
  in wireless communication.
\newblock {\em science}, 304(5667):78--80, 2004.

\bibitem{salmen2005echo}
Matthias Salmen and Paul~G Ploger.
\newblock Echo state networks used for motor control.
\newblock In {\em Proceedings of the 2005 IEEE international conference on
  robotics and automation}, pages 1953--1958. IEEE, 2005.

\bibitem{tong2007learning}
Matthew~H Tong, Adam~D Bickett, Eric~M Christiansen, and Garrison~W Cottrell.
\newblock Learning grammatical structure with echo state networks.
\newblock {\em Neural networks}, 20(3):424--432, 2007.

\bibitem{gallicchio2018comparison}
Claudio Gallicchio, Alessio Micheli, and Luca Pedrelli.
\newblock Comparison between deepesns and gated rnns on multivariate
  time-series prediction.
\newblock {\em arXiv preprint arXiv:1812.11527}, 2018.

\bibitem{vlachas2020backpropagation}
Pantelis~R Vlachas, Jaideep Pathak, Brian~R Hunt, Themistoklis~P Sapsis,
  Michelle Girvan, Edward Ott, and Petros Koumoutsakos.
\newblock Backpropagation algorithms and reservoir computing in recurrent
  neural networks for the forecasting of complex spatiotemporal dynamics.
\newblock {\em Neural Networks}, 2020.

\bibitem{popov2019echo}
Alexander Popov, Petia Koprinkova-Hristova, Kiril Simov, and Petya Osenova.
\newblock Echo state vs. lstm networks for word sense disambiguation.
\newblock In {\em International Conference on Artificial Neural Networks},
  pages 94--109. Springer, 2019.

\bibitem{jirak2020echo}
Doreen Jirak, Stephan Tietz, Hassan Ali, and Stefan Wermter.
\newblock Echo state networks and long short-term memory for continuous gesture
  recognition: a comparative study.
\newblock {\em Cognitive Computation}, pages 1--13, 2020.

\bibitem{karpathy2015visualizing}
Andrej Karpathy, Justin Johnson, and Li~Fei-Fei.
\newblock Visualizing and understanding recurrent networks.
\newblock {\em arXiv preprint arXiv:1506.02078}, 2015.

\bibitem{DBLP:journals/corr/RadfordJS17}
Alec Radford, Rafal J{\'{o}}zefowicz, and Ilya Sutskever.
\newblock Learning to generate reviews and discovering sentiment.
\newblock {\em CoRR}, abs/1704.01444, 2017.

\bibitem{strobelt2017lstmvis}
Hendrik Strobelt, Sebastian Gehrmann, Hanspeter Pfister, and Alexander~M Rush.
\newblock Lstmvis: A tool for visual analysis of hidden state dynamics in
  recurrent neural networks.
\newblock {\em IEEE transactions on visualization and computer graphics},
  24(1):667--676, 2017.

\bibitem{madsen2019visualizing}
Andreas Madsen.
\newblock Visualizing memorization in rnns.
\newblock {\em Distill}, 4(3):e16, 2019.

\bibitem{bianchi2016investigating}
Filippo~Maria Bianchi, Lorenzo Livi, and Cesare Alippi.
\newblock Investigating echo-state networks dynamics by means of recurrence
  analysis.
\newblock {\em IEEE transactions on neural networks and learning systems},
  29(2):427--439, 2016.

\bibitem{Jaeger2014}
Herbert Jaeger.
\newblock Controlling recurrent neural networks by conceptors.
\newblock {\em arXiv preprint arXiv:1403.3369}, 2014.

\bibitem{Jaeger2017}
Herbert Jaeger.
\newblock Using conceptors to manage neural long-term memories for temporal
  patterns.
\newblock {\em Journal of Machine Learning Research}, 18(13):1--43, 2017.

\bibitem{mcinnes2018umap}
Leland McInnes, John Healy, and James Melville.
\newblock Umap: Uniform manifold approximation and projection for dimension
  reduction.
\newblock {\em arXiv preprint arXiv:1802.03426}, 2018.

\bibitem{carter2019activation}
Shan Carter, Zan Armstrong, Ludwig Schubert, Ian Johnson, and Chris Olah.
\newblock Activation atlas.
\newblock {\em Distill}, 4(3):e15, 2019.

\bibitem{devlin2018bert}
Jacob Devlin, Ming-Wei Chang, Kenton Lee, and Kristina Toutanova.
\newblock Bert: Pre-training of deep bidirectional transformers for language
  understanding.
\newblock {\em arXiv preprint arXiv:1810.04805}, 2018.

\bibitem{coenen2019visualizing}
Andy Coenen, Emily Reif, Ann Yuan, Been Kim, Adam Pearce, Fernanda Vi{\'e}gas,
  and Martin Wattenberg.
\newblock Visualizing and measuring the geometry of bert.
\newblock {\em arXiv preprint arXiv:1906.02715}, 2019.

\bibitem{boggust2019embedding}
Angie Boggust, Brandon Carter, and Arvind Satyanarayan.
\newblock Embedding comparator: Visualizing differences in global structure and
  local neighborhoods via small multiples, 2019.

\bibitem{sign-language-lstm}
Visualizing lstm networks, australian sign language model visualization.
\newblock
  \url{https://medium.com/asap-report/visualizing-lstm-networks-part-i-f1d3fa6aace7}.
\newblock Accessed: 2020-11-25.

\bibitem{abdelrahman2019analyzing}
Mohamed Mahmoud Hafez~Mahmoud Abdelrahman.
\newblock {\em Analyzing robustness of models of chaotic dynamical systems
  learned from data with Echo state networks}.
\newblock PhD thesis, Rice University, 2019.

\bibitem{juven2020cross}
Alexis Juven and Xavier Hinaut.
\newblock Cross-situational learning with reservoir computing for language
  acquisition modelling.
\newblock In {\em 2020 International Joint Conference on Neural Networks (IJCNN
  2020)}, 2020.

\bibitem{dominey2006neurolinguistic}
Peter~Ford Dominey, Michel Hoen, and Toshio Inui.
\newblock A neurolinguistic model of grammatical construction processing.
\newblock {\em Journal of Cognitive Neuroscience}, 18(12):2088--2107, 2006.

\bibitem{hinaut2013real}
Xavier Hinaut and Peter~Ford Dominey.
\newblock Real-time parallel processing of grammatical structure in the
  fronto-striatal system: A recurrent network simulation study using reservoir
  computing.
\newblock {\em PloS one}, 8(2):e52946, 2013.

\bibitem{hinaut2015recurrent}
Xavier Hinaut, Johannes Twiefel, Maxime Petit, Peter Dominey, and Stefan
  Wermter.
\newblock A recurrent neural network for multiple language acquisition:
  Starting with english and french.
\newblock In {\em Proceedings of the NIPS Workshop on Cognitive Computation:
  Integrating Neural and Symbolic Approaches (CoCo 2015)}, 2015.

\bibitem{hinaut2019teach}
Xavier Hinaut and Johannes Twiefel.
\newblock Teach your robot your language! trainable neural parser for modelling
  human sentence processing: Examples for 15 languages.
\newblock {\em IEEE Transactions on Cognitive and Developmental Systems}, 2019.

\bibitem{hinaut2014exploring}
Xavier Hinaut, Maxime Petit, Gregoire Pointeau, and Peter~Ford Dominey.
\newblock Exploring the acquisition and production of grammatical constructions
  through human-robot interaction with echo state networks.
\newblock {\em Frontiers in neurorobotics}, 8:16, 2014.

\bibitem{twiefel2016semantic}
Johannes Twiefel, Xavier Hinaut, and Stefan Wermter.
\newblock Semantic role labelling for robot instructions using echo state
  networks.
\newblock 2016.

\bibitem{hinaut2018input}
Xavier Hinaut.
\newblock Which input abstraction is better for a robot syntax acquisition
  model? phonemes, words or grammatical constructions?
\newblock In {\em 2018 Joint IEEE 8th International Conference on Development
  and Learning and Epigenetic Robotics (ICDL-EpiRob)}, pages 281--286. IEEE,
  2018.

\bibitem{zhong2017toward}
Junpei Zhong, Angelo Cangelosi, and Tetsuya Ogata.
\newblock Toward abstraction from multi-modal data: empirical studies on
  multiple time-scale recurrent models.
\newblock In {\em 2017 International Joint Conference on Neural Networks
  (IJCNN)}, pages 3625--3632. IEEE, 2017.

\bibitem{ororbia2018like}
Alexander~G Ororbia, Ankur Mali, Matthew~A Kelly, and David Reitter.
\newblock Like a baby: Visually situated neural language acquisition.
\newblock {\em arXiv preprint arXiv:1805.11546}, 2018.

\bibitem{Jaeger2001}
Herbert Jaeger.
\newblock The "echo state" approach to analysing and training recurrent neural
  networks.
\newblock Technical Report 148, German National Research Center for Information
  Technology GMD, Bonn, Germany, 1 2001.

\bibitem{Lukoeviius2009}
Mantas Luko{\v{s}}evi{\v{c}}ius and Herbert Jaeger.
\newblock Reservoir computing approaches to recurrent neural network training.
\newblock {\em Computer Science Review}, 3(3):127--149, 8 2009.

\bibitem{trouvain2020reservoirpy}
Nathan Trouvain, Luca Pedrelli, Thanh~Trung Dinh, and Xavier Hinaut.
\newblock Reservoirpy: an efficient and user-friendly library to design echo
  state networks.
\newblock 2020.

\bibitem{sussillo2009generating}
David Sussillo and Larry~F Abbott.
\newblock Generating coherent patterns of activity from chaotic neural
  networks.
\newblock {\em Neuron}, 63(4):544--557, 2009.

\bibitem{kingma2014adam}
Diederik~P Kingma and Jimmy Ba.
\newblock Adam: A method for stochastic optimization.
\newblock {\em arXiv preprint arXiv:1412.6980}, 2014.

\bibitem{rigotti2013importance}
Mattia Rigotti, Omri Barak, Melissa~R Warden, Xiao-Jing Wang, Nathaniel~D Daw,
  Earl~K Miller, and Stefano Fusi.
\newblock The importance of mixed selectivity in complex cognitive tasks.
\newblock {\em Nature}, 497(7451):585--590, 2013.

\bibitem{enel2016reservoir}
Pierre Enel, Emmanuel Procyk, Ren{\'e} Quilodran, and Peter~Ford Dominey.
\newblock Reservoir computing properties of neural dynamics in prefrontal
  cortex.
\newblock {\em PLoS computational biology}, 12(6):e1004967, 2016.

\bibitem{maaten2008visualizing}
Laurens van~der Maaten and Geoffrey Hinton.
\newblock Visualizing data using t-sne.
\newblock {\em Journal of machine learning research}, 9(Nov):2579--2605, 2008.

\bibitem{carmantini2017modular}
Giovanni~S Carmantini, Peter Beim~Graben, Mathieu Desroches, and Serafim
  Rodrigues.
\newblock A modular architecture for transparent computation in recurrent
  neural networks.
\newblock {\em Neural Networks}, 85:85--105, 2017.

\bibitem{tabor2003learning}
Whitney Tabor.
\newblock Learning exponential state-growth languages by hill climbing.
\newblock {\em IEEE Transactions on Neural Networks}, 14(2):444--446, 2003.

\bibitem{tabor2000fractal}
Whitney Tabor.
\newblock Fractal encoding of context-free grammars in connectionist networks.
\newblock {\em Expert Systems}, 17(1):41--56, 2000.

\bibitem{understanding-lstm-embedding}
Understanding neural networks by embedding hidden representations.
\newblock \url{https://rakeshchada.github.io/Neural-Embedding-Animation.html}.
\newblock Accessed: 2020-11-25.

\bibitem{verstraeten2010memory}
David Verstraeten, Joni Dambre, Xavier Dutoit, and Benjamin Schrauwen.
\newblock Memory versus non-linearity in reservoirs.
\newblock In {\em The 2010 international joint conference on neural networks
  (IJCNN)}, pages 1--8. IEEE, 2010.

\bibitem{lukovsevivcius2012practical}
Mantas Luko{\v{s}}evi{\v{c}}ius.
\newblock A practical guide to applying echo state networks.
\newblock In {\em Neural networks: Tricks of the trade}, pages 659--686.
  Springer, 2012.

\bibitem{pascanu2011neurodynamical}
Razvan Pascanu and Herbert Jaeger.
\newblock A neurodynamical model for working memory.
\newblock {\em Neural networks}, 24(2):199--207, 2011.

\bibitem{hoerzer2014emergence}
Gregor~M Hoerzer, Robert Legenstein, and Wolfgang Maass.
\newblock Emergence of complex computational structures from chaotic neural
  networks through reward-modulated hebbian learning.
\newblock {\em Cerebral cortex}, 24(3):677--690, 2014.

\bibitem{strock2020robust}
Anthony Strock, Xavier Hinaut, and Nicolas~P Rougier.
\newblock A robust model of gated working memory.
\newblock {\em Neural Computation}, 32(1):153--181, 2020.

\end{thebibliography}

\section{Supplementary Material}
\label{sec-sup-mat}

\subsection{Influence of the choice of the threshold factor}
\label{subsec-threshold_factor_influence}

\begin{figure}
    \centering
        \begin{subfigure}{\textwidth}
        \centering
         \includegraphics[width=0.9\textwidth]{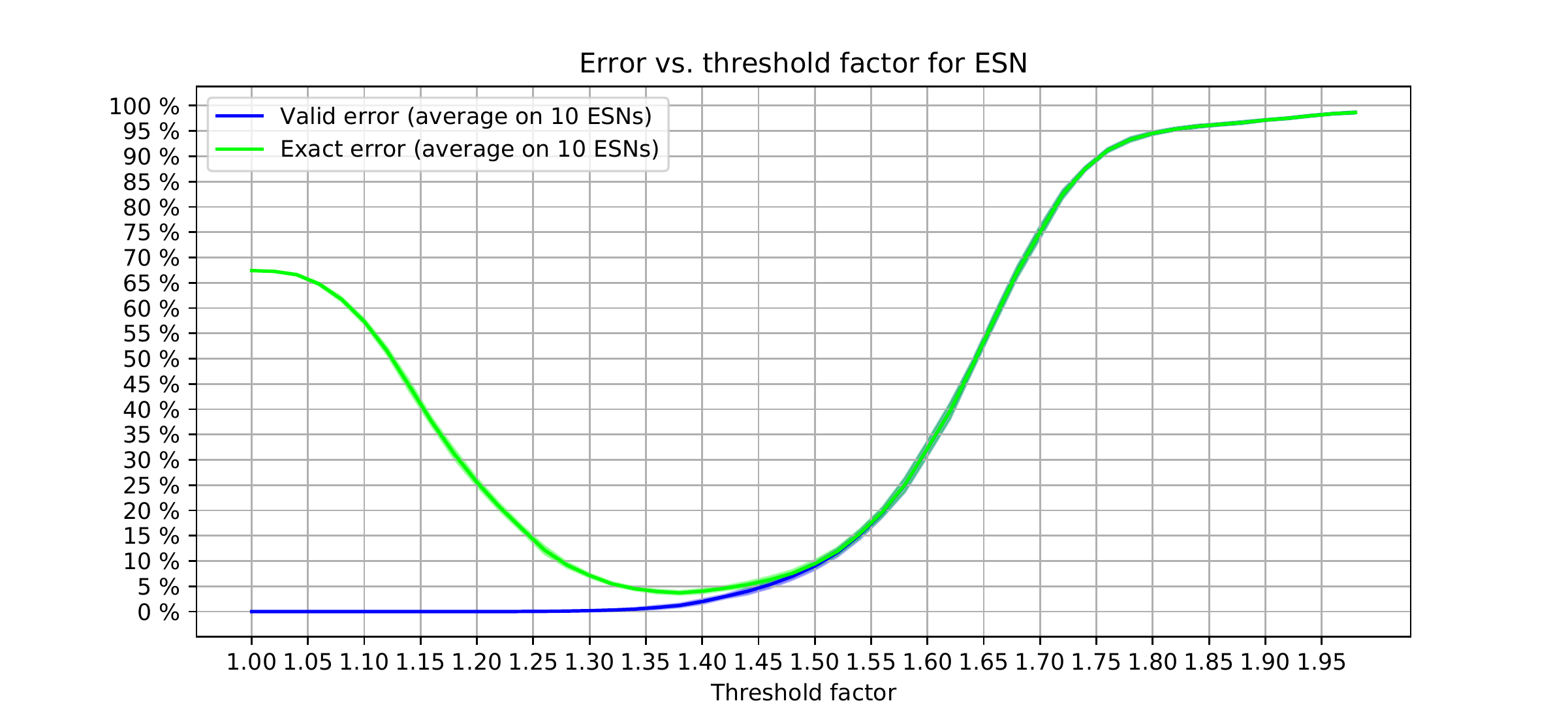}
         \caption{Influence of the threshold factor for the ESN.}
    \end{subfigure}
        \begin{subfigure}{\textwidth}
        \centering
         \includegraphics[width=0.9\textwidth]{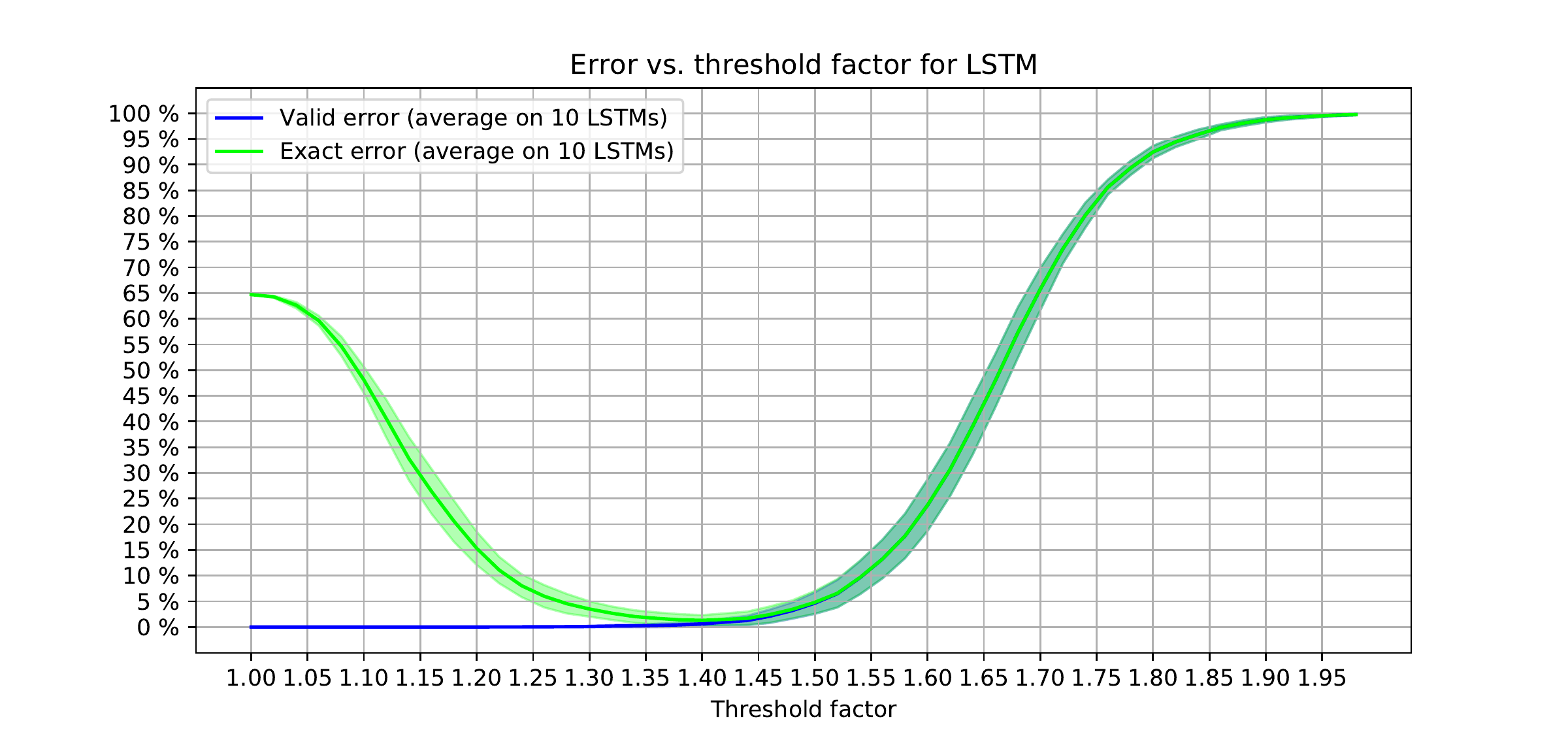}
         \caption{Influence of the threshold factor for the LSTM.}
    \end{subfigure}
    \caption{Influence of the threshold factor on the errors for the LSTM and the ESN. Results are averaged on 10 independent instances. The width of the filled area around the curves is the standard deviation for the 10 instances. We can see that the effect are similar for both models. The fixed choice of 1.3 done in this paper doesn't favour one model over the other.}
    \label{fig:effect_threshold_fact}
\end{figure}

We investigated the influence of the threshold factor on performances. We can see in figure \ref{fig:effect_threshold_fact} that the higher the threshold factor, the lesser the exact error but the higher the valid error. In fact, the role of the exact error is similar to false positive and the valid error to false negative. That's why we observe this trade-off between the two error type. There is an optimal value for the exact error between the extreme cases when no characteristics are given to the objects (high threshold) and when we chose every maximal activated characteristic in the final representation (low threshold). 

In figure \ref{fig:effect_threshold_fact}, we can see that the global shape curve of the LSTM and the ESN are similar: the minimum is attained around the same value, 1.35. So by choosing the threshold factor equals to 1.3, we aren't advantaging one model over the other. In fact, we can consider this choice of threshold factor as a part of the task definition: we fixed the process to transform the output vector of the model into a discrete representation.

\subsection{Output activity analysis on a sentence structure outside of the data space}

\begin{figure}
    \centering
    \includegraphics[width = \textwidth]{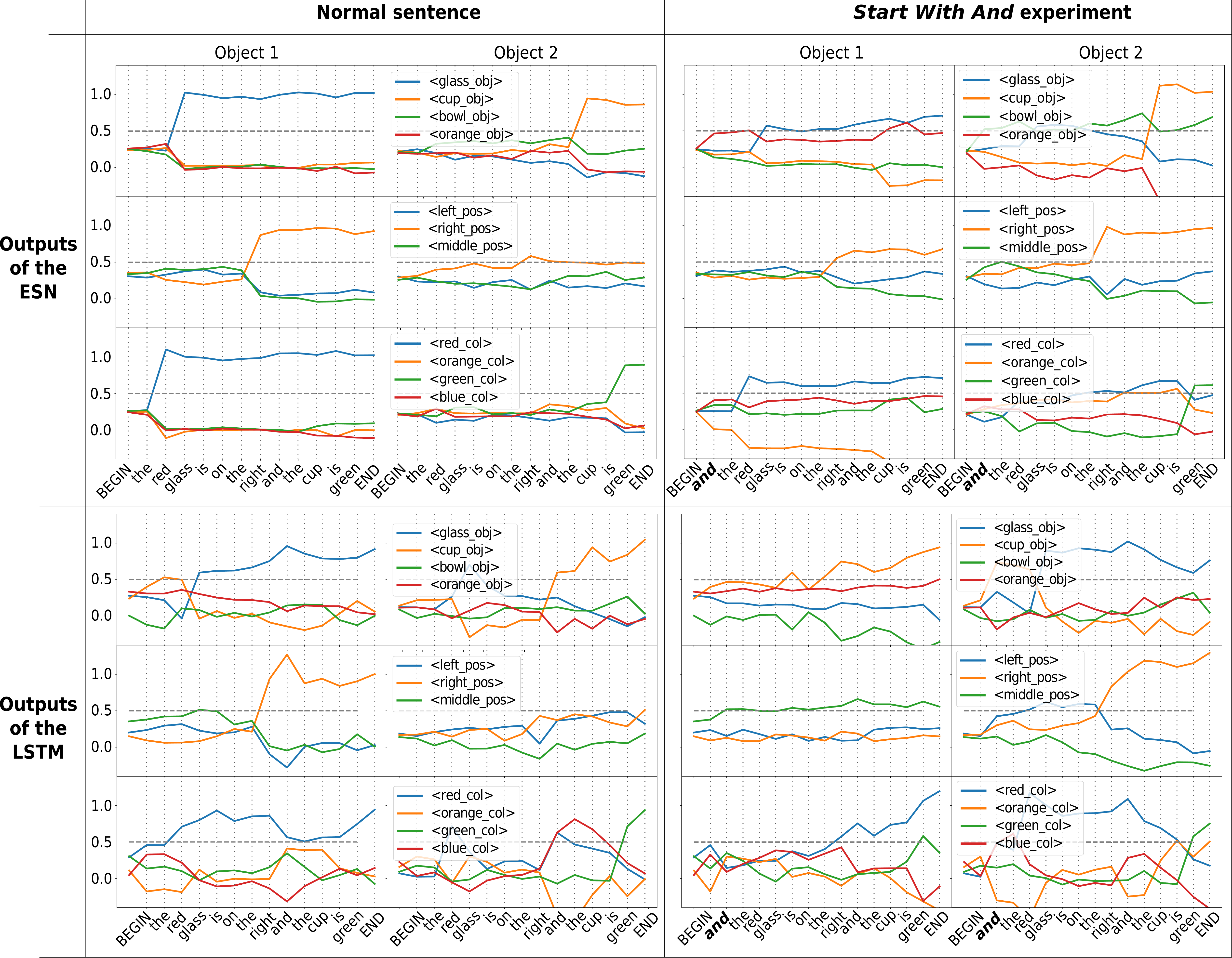}
    \caption{Comparison of activation plot for a normal sentence and for a sentence beginning with the word \enquote{and}. We can see that the word \enquote{and} seems to trigger an inner switch that controls to which object correspond the described characteristics. It is especially visible for the concept relative to the position of the objects. Interestingly, even if the architecture of LSTM and ESN are really different, they both react in the same way, but to different extent, to this experiment.}
    \label{fig:SWA_plots}
\end{figure}

Beyond visualizing activation plot for sentences from the dataset as we have done in \ref{comparing_output_activation}, we can also see how both models react to a sentence with a structure they have never seen. With this idea in mind, we crafted an experiment to study how the models learnt that the sentences are divided in two part: the description of the first object and then the second. We expected that the word \enquote{and} played an important role in this division. To test this hypothesis we created the experiment Start With And (SWA) where we observed the reaction of the model to a sentence beginning with the word \enquote{and}. The results in figure \ref{fig:SWA_plots} show that with the word \enquote{and} at the beginning of the sentence, the activation of the first object seems inhibited and the outputs of the second object inherit some of the characteristics of the first. The effect is clearer for LSTM than for the ESN, nonetheless it is still observable for both. This experiment suggests that there is an inner switch that triggers when the word \enquote{and} is seen. When the switch is off, the keyword corresponding to a characteristic activate the first object and after the switch turned on, the words will activate primarily the concepts related to the second object. This hypothesis is supported by our finding presented in section \ref{sec-understand_hidden_cells} where we identified reservoir units and LSTM cell reacting specifically to the word \enquote{and}.

\subsection{The winglet effect}

\begin{figure}
    \centering
    \includegraphics[width = \textwidth]{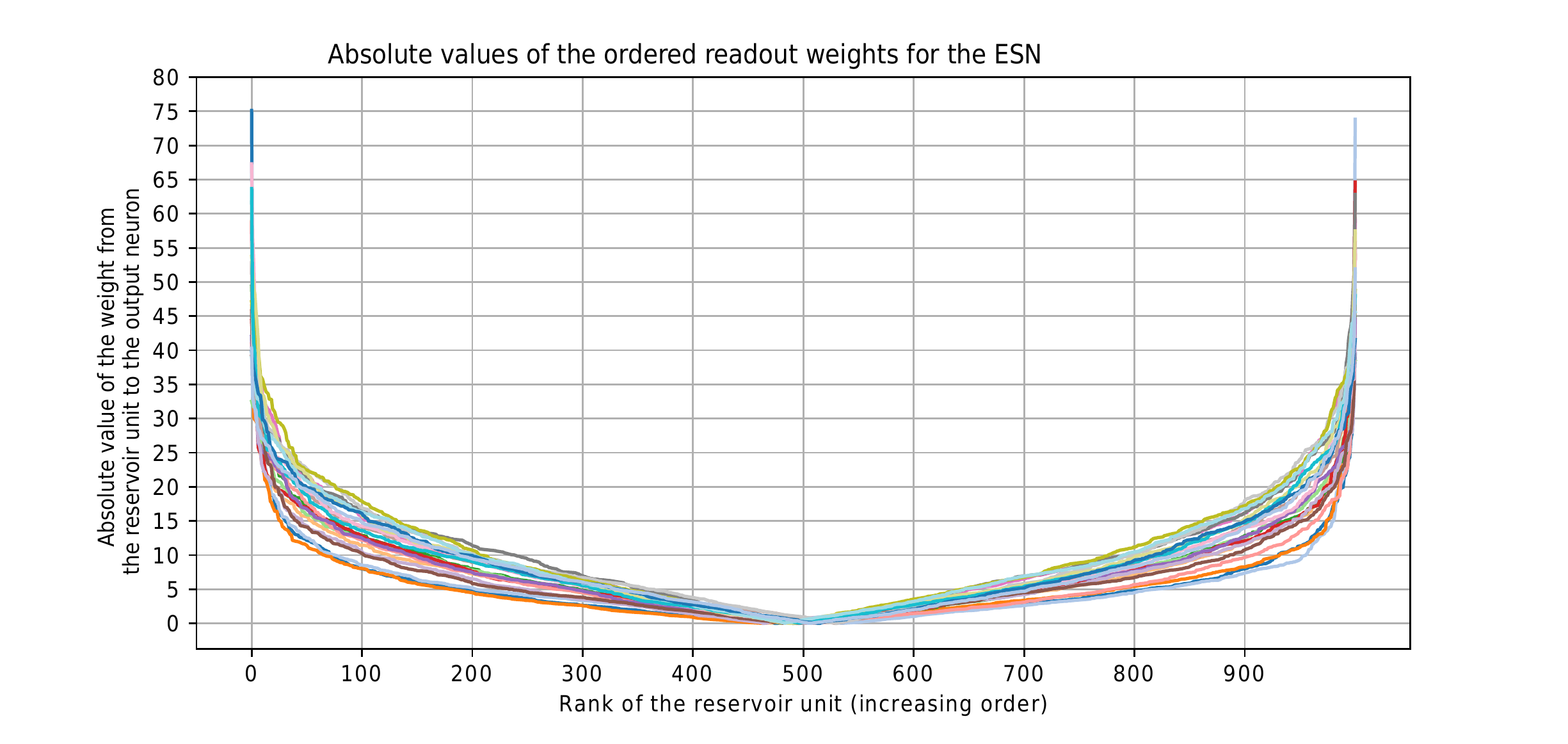}
    \caption{We plotted the ordered absolute values of the connection weight between the reservoir unit and each output neuron. Each line correspond to an output neuron. The connection are ordered in increasing order: this means that negative values are on the left and the positive ones on the right. Because the reservoir units activation are randomly ordered around zero, there are as much positive as negative weights. Due to regularization penalizing high weight values, only a little number of reservoir units gather high weight values. For a given output neuron, it seems to be a small group of very useful reservoir units. Indeed, 8\% of the readout matrix weights account for 28\% in the total absolute sum. The ESN used here was trained with FL.}
    \label{fig:winglet_effect}
\end{figure}

When we conducted the selection of the reservoir units the most connected to a given output neuron in section \ref{analysis_of_reservoir_units_activation}, we also analyse the distribution of the weights in the ESN readout matrix. For a given output neurons, there is 1,001 weights: one for each reservoir units and one for the constant units. We sorted these 1,001 weights and then plotted their absolute values in order to be able to easily compare negative and positive values. We did this for all the 22 output neurons of the ESN. The results are shown in figure \ref{fig:winglet_effect}.
Because the FORCE learning algorithm uses regularisation, high weights values are penalized. This means that high weights are kept only if they are really useful to minimize the error. Here we can see that the weight distribution is highly heterogeneous: a small number of reservoir units gather a great portion of the weights connections values. For a given output neuron, it seems to exist a small group of reservoir units highly useful for predicting the output. These are the ones investigated in section \ref{analysis_of_reservoir_units_activation}. This analysis can be seen as a form of feature selection to understand the inner working of the ESN.

Moreover, the figure \ref{fig:winglet_effect} is symmetrical. For a given output neuron, it seems to be as much connection to the reservoir with a positive weight than with negative weight. This can be explained by the fact that the reservoir units activation are randomly distributed around zero. So for any targeted output, the weights will also be distributed symmetrically around zero. This shape reminds us of the \emph{winglet}, this small curved surface placed at the end of plane wings. That's why we called this phenomena the \emph{winglet effect}. %

\subsection{Source code}
The code of the implementation of the task and the models used, as well as the code used for the analysis of the hidden cells and the creation of RSSviz is available here: \href{https://github.com/aVariengien/esn-vs-lstm-on-cross-situationnal-learning}{\texttt{https://github.com/aVariengien/esn-vs-lstm-on-cross-situationnal-learning}}.

\end{document}